\documentclass{article}

\PassOptionsToPackage{numbers,sort&compress}{natbib}
\PassOptionsToPackage{dvipsnames}{xcolor}
\usepackage{arxiv-ellis}

\usepackage{nicefrac}
\usepackage{arydshln}
\usepackage{bbm}
\usepackage{multirow}
\usepackage{tcolorbox}
\usepackage{subfig}
\usepackage{tikz}
\usepackage{chemfig}
\usepackage[version=4]{mhchem}
\tcbuselibrary{skins,breakable}
\usepackage{xurl}
\usepackage{breakurl}
\usepackage[export]{adjustbox}
\usepackage{wrapfig}
\usepackage{array}
\usepackage{chemformula}
\usepackage{algorithm}
\usepackage{algorithmic}
\usepackage{amsthm}

\theoremstyle{plain}
\newtheorem{theorem}{Theorem}[section]

\theoremstyle{definition}
\newtheorem{definition}[theorem]{Definition}

\theoremstyle{remark}
\newtheorem{remark}[theorem]{Remark}

\usepackage{amsmath,amsfonts,bm}

\def\eqref#1{equation~\ref{#1}}

\def\1{\bm{1}}

\DeclareMathAlphabet{\mathsfit}{\encodingdefault}{\sfdefault}{m}{sl}
\SetMathAlphabet{\mathsfit}{bold}{\encodingdefault}{\sfdefault}{bx}{n}

\newcommand{\eqdef}{\stackrel{\text{def}}{=}}

\newcommand{\vect}[1]{\boldsymbol{#1}}
\newcommand{\ie}{{\em i.e.\/}}

\newcommand{\Pcal}{{\mathcal P}}

\newcommand{\Fcal}{{\mathcal F}}

\newcommand{\Ocal}{{\mathcal O}}

\newcommand{\Xcal}{{\mathcal X}}

\newcommand{\Reals}{{\mathbb{R}}}

\title{Constrained Look-ahead Guidance for \\ Interference-Aware Flow Editing}

\author{%
  \textbf{Doudou Zhang}\textsuperscript{1,2}\quad
  \textbf{Qi Chen}\textsuperscript{1,2}\thanks{Corresponding author.}\\[7pt]
  {\small
    \textsuperscript{1}ELLIS Institute Finland\quad
    \textsuperscript{2}Aalto University
  }\\[5pt]
  {\small
    \texttt{doudou.zhang@aalto.fi}\quad
    \texttt{qi.chen.inf@aalto.fi}
  }
}
\hypersetup{
  pdftitle={Constrained Look-ahead Guidance for Interference-Aware Flow Editing},
  pdfauthor={Doudou Zhang, Qi Chen}
}

\begin{document}

\maketitle

\begin{abstract}
Diffusion and flow models enable powerful generative modeling, but controlled editing remains challenging due to the need to balance source fidelity and edit compliance along the generative trajectory. We propose a principled, inversion-free framework for flow-based image editing from a trajectory centric perspective. Rather than relying on inversion or heuristic guidance, we introduce a one-step look-ahead mechanism that evaluates the effect of candidate updates on both edit and fidelity rewards estimated via denoising, and adjusts the current update based on this evaluation. At each step, our method applies reward-driven corrections guided by the look-ahead evaluation under an explicit fidelity constraint, resulting in stable and coherent trajectory evolution. This look-ahead design mitigates gradient interference between edit and fidelity objectives by anticipating their interaction in future states, leading to improved trade-offs. Experiments demonstrate that the proposed approach achieves stronger semantic edits while better preserving source content compared to existing strong baselines.
\tcblower
{\sffamily\bfseries Code:} \url{https://github.com/aalto-icl/ConstrainedLookaheadGuidance}
\end{abstract}

\section{Introduction}\label{sec:intro}

Diffusion~\cite{song2019generative, song2020score,ho2020denoising} and flow models~\cite{lipman2022flow,liuFlowStraightFast2022} have achieved remarkable success in generative modeling~\cite{chenGeneralizationVAEDiffusion2024} by transforming noise into structured outputs through iterative denoising, enabling applications such as image synthesis~\cite{rombach2022high,esserScalingRectifiedFlow2024} and video generation~\cite{peebles2023scalable}. With the proliferation of large-scale pretrained text–image models~\cite{rombach2022high, labs2025flux}, there is increasing interest in leveraging them for test-time, text-guided image editing~\cite{kulikovFlowEditInversionFreeTextBased2025, hanImprovingTuningFreeReal2023, huberman-spiegelglasEditFriendlyDDPM2024}, where semantic changes are specified via natural language.

Despite rapid progress, the core challenge of text-guided image editing remains insufficiently understood from a principled standpoint. At its core, editing requires transforming a given sample along a generative trajectory to satisfy two competing objectives: preserving fidelity to the source image while aligning with the target semantics. This naturally motivates a \emph{multi-objective optimization view}. However, existing methods rarely adopt this perspective explicitly or provide principled mechanisms to enforce such trade-offs, often resulting in heuristic and unstable behavior in practice. 

Finetuning-based methods~\cite{ruiz2023dreambooth,han2023svdiff,kumari2023multi,chen2023disenbooth} achieve strong fidelity and personalization by adapting model parameters, but incur substantial training overhead and reduced flexibility, as adapting to new concepts or instances generally requires additional optimization. Inversion-based approaches~\cite{miyakeNegativepromptInversionFast2024,mokadyNULLTextInversionEditing2023,peebles2023scalable}, such as DDIM inversion \cite{songDenoisingDiffusionImplicit2020}, map an input image to a latent noise code and reconstruct it through the reverse process. By injecting editing conditions (e.g., text prompts) into this latent, the sampling trajectory can be guided toward target semantics. However, this inversion-based paradigm often yields limited fidelity to the source image~\cite{huberman-spiegelglasEditFriendlyDDPM2024}, primarily due to inaccurate inversion and error accumulation across timesteps, which produce noise trajectories that are not well-suited for controlled editing. To address these limitations, subsequent methods aim to improve inversion accuracy~\cite{hanImprovingTuningFreeReal2023,miyakeNegativepromptInversionFast2024,peebles2023scalable} or modify the sampling process, for example by injecting structural cues such as attention maps extracted during inversion~\cite{hertzPrompttoPromptImageEditing2022,tumanyanPlugandPlayDiffusionFeatures2023,caoMasaCtrlTuningFreeMutual2023} or by incorporating test-time optimization~\cite{mokadyNULLTextInversionEditing2023,kawarImagicTextBasedReal2023} to enforce more faithful reconstruction. More recent inversion-free approaches, including FlowEdit~\cite{kulikovFlowEditInversionFreeTextBased2025}, bypass inversion by directly modifying the generative vector field. While more efficient, they typically rely on heuristic updates without explicit objectives, which can result in weak or unstable edits. 

In this work, we formulate text-guided image editing as a \emph{constrained multi-objective optimization} problem over the generative trajectory. Typical requirement to preserve source fidelity naturally induces a constrained formulation, where edit alignment is optimized subject to such fidelity constraints. Rather than modifying the model or relying on heuristic vector field differences, we perform a local constrained update at each timestep to guide the trajectory. Our key idea is to estimate edit and fidelity rewards via one-step denoising and introduce a \emph{look-ahead} mechanism 
that evaluates these rewards at the predicted clean endpoints following simulated objective-specific updates. The resulting update incorporates post-intervention reward information to better navigate the edit–fidelity trade-off, while a bounded fidelity correction is applied when necessary to control constraint violations. By anticipating how updates affect future states, this design implicitly resolves the multi-objective conflict and reduces gradient interference between edit and fidelity objectives.

This leads to a well-controlled, inversion-free editing process that achieves strong semantic edits while preserving source content. Our key insight is that trajectory-level control can be realized through a sequence of principled per-step constrained optimizations, rather than global optimization or heuristic guidance. Specifically, our contributions are as follows:

\textbf{Constrained trajectory-level editing formulation.}
We formulate text-guided flow editing as a constrained multi-objective control
problem over the generative trajectory, where edit compliance is optimized only
within a fidelity-feasible region.

\textbf{Look-ahead surrogate for local Pareto control.}
Since directly solving the constrained Pareto update at every denoising step is computationally challenging, we derive a one-step look-ahead surrogate that evaluates edit and fidelity rewards at the predicted clean endpoints following simulated objective-specific updates. This yields a locally non-myopic update direction that better navigates the edit–fidelity trade-off by accounting for post-update reward information.

\textbf{Interference-aware editing behavior.}
We empirically show that the proposed update improves edit-fidelity gradient
alignment, reduces violations of the fidelity constraint, and yields a more
favorable edit-fidelity trade-off than greedy edit-first updates. Experiments further demonstrate that our approach enables stronger semantic edits with better source
preservation than strong baselines.

\section{Related Work}

\paragraph{Finetuned-based Editing.} Recent works address text-to-image (T2I) generation by fine-tuning pre-trained diffusion or flow models on a few personalized images. Methods such as DreamBooth~\cite{ruiz2023dreambooth} perform full-model fine-tuning to bind a subject to a unique token, achieving strong identity preservation but requiring costly optimization. DiffusionCLIP~\cite{kimDiffusionclipTextguidedDiffusion2022} performs instance-specific fine-tuning guided by CLIP objectives. Subsequent works improve scalability via multi-concept customization~\cite{kumari2023multi} or disentangled identity modeling~\cite{chen2023disenbooth}. To reduce training cost, parameter-efficient approaches such as LoRA~\cite{hu2022lora} were adopted, where SVDiff~\cite{han2023svdiff}, and T2I-Adapter~\cite{mou2024t2i} restrict updates to low-rank or modular components. Alternatively, textual inversion methods~\cite{gal2022image,daras2022multiresolution,voynov2023p+} optimize token embeddings without modifying model weights, offering lightweight adaptation but limited expressiveness. Despite strong performance, these approaches typically require non-trivial optimization or careful tuning at test time, limiting their practicality for efficient editing.

\textbf{Inversion-based Editing.}~
Inversion-based methods first map the source image to a noisy latent representation, from which edited samples are generated to maintain consistency with the original input~\cite{songDenoisingDiffusionImplicit2020,staniszewskiThereBackAgain2026}. However, naive inversion often results in insufficient reconstruction fidelity~\cite{huberman-spiegelglasEditFriendlyDDPM2024}. While several approaches have been proposed to improve inversion accuracy \cite{hanImprovingTuningFreeReal2023,miyakeNegativepromptInversionFast2024,mokadyNULLTextInversionEditing2023,wallace2023edict}, even near-exact inversion on synthetic data does not fully resolve this issue~\cite{huberman-spiegelglasEditFriendlyDDPM2024}.
To address these limitations, a line of work instead intervenes directly in the sampling process by leveraging structural information extracted during inversion and encoded within the model. Attention-control methods~\cite{hertzPrompttoPromptImageEditing2022,caoMasaCtrlTuningFreeMutual2023,tumanyanPlugandPlayDiffusionFeatures2023,zhuKVEditTrainingFreeImage2025,shinExploringMultimodalDiffusion2025,wangTamingRectifiedFlow2024} preserve spatial layout by manipulating or reusing cross-attention or self-attention maps during generation.
However, these methods are often tightly coupled to specific model architectures or sampling procedures, which limits their adaptability to new settings. Another line of work aims to conduct \emph{test-time optimization} to improve either source fidelity or edit compliance~\cite{yuFreeDoMTrainingFreeEnergyGuided2023, patelSteeringRectifiedFlow2024, nguyenHEditEffectiveFlexible2025,changTrainingFreeRewardGuidedImage2026}. Null-text Inversion~\cite{mokadyNULLTextInversionEditing2023} explicitly optimizes the unconditional text embedding during inversion to improve reconstruction. To improve edit compliance, Imagic~\citep{kawarImagicTextBasedReal2023} optimizes text embeddings and fine-tunes the diffusion model, while Pix2Pix-Zero~\cite{parmarZeroshotImagetoImageTranslation2023} estimates an editing direction in the text embedding space, which can be viewed as an implicit or amortized form of optimization. These approaches highlight the effectiveness of leveraging differentiable objectives at test time. Other methods such as Negative-prompt Inversion~\cite{miyakeNegativepromptInversionFast2024}, Direct Inversion~\cite{juDirectInversionBoosting2023}, and~\cite{hanImprovingTuningFreeReal2023,garibiReNoiseRealImage2024,routSemanticImageInversion2024,mengSDEditGuidedImage2021} modify the inversion or sampling trajectory to obtain source-consistent latents without explicit optimization. 

\textbf{Inversion-free Editing.}~Trajectory guidance methods modify intermediate sampling states using gradient-based updates derived from differentiable objectives, enabling controllable generation without requiring inversion~\cite{yuFreeDoMTrainingFreeEnergyGuided2023,bansalUniversalGuidanceDiffusion2023a,patelSteeringRectifiedFlow2024,xuInversionFreeImageEditing2023}. However, as they generate samples from noise rather than reconstructing source-specific latent representations, they typically lack explicit mechanisms to enforce source fidelity.
In contrast, FlowEdit~\cite{kulikovFlowEditInversionFreeTextBased2025} performs inversion-free editing by constructing an ODE that directly maps between the source and target text conditioned distributions, avoiding explicit latent reconstruction. 

Unlike approaches mentioned above that treat editing and fidelity separately, we model edit compliance and source fidelity as \emph{coupled, differentiable objectives} along the denoising trajectory, yielding a multi-objective optimization view. As both objectives share the same trajectory, naive aggregation induces gradient conflicts. Building on classical multi-objective optimization methods that mitigate gradient conflicts through projection or Pareto-based updates~\cite{desideriMultiplegradientDescentAlgorithm2012,yuGradientSurgeryMultiTask2020,liuConflictaverseGradientDescent2021}, recent image editing methods have explored strategies to balance editability and fidelity \cite{wangRethinkingRealImage2026,maoTuningFreeImageEditing2025}.However, these approaches primarily focus on balancing competing objectives without explicitly accounting for the downstream effects of individual updates. In contrast, we introduce a look-ahead formulation that enables non-myopic, per-step constrained optimization, with explicit fidelity constraints for precise editing control.

\section{Problem Formulation}\label{sec:formalization}

Suppose we are given a source content $x^{\mathrm{src}}$ and a condition $c^{\mathrm{src}}$ describing it. Our goal is to edit $x^{\mathrm{src}}$ according to a desired condition $c^{\mathrm{edit}}$. In image editing, $x^{\mathrm{src}}$ denotes the latent of the source image, and conditions are typically text prompts. A common approach first inverts $x^{\mathrm{src}}$ to a noisy latent via
\begin{equation}
\label{eq:inv_ode}
d x_t^{\mathrm{inv}} = v_\theta(x_t^{\mathrm{inv}}, t, c^{\mathrm{src}})\,dt,\quad 
t \in [t_0, t_T],\quad x_{t_0}^{\mathrm{inv}} = x^{\mathrm{src}},
\end{equation}
yielding $x_{t_T}^{\mathrm{inv}}$, and then generates the edited result via
\begin{equation}
\label{eq:gen_ode}
d x_t^{\mathrm{gen}} = v_\theta(x_t^{\mathrm{gen}}, t, c^{\mathrm{edit}})\,dt,\quad 
t \in [t_T, t_0],\quad x_{t_T}^{\mathrm{gen}} = x_{t_T}^{\mathrm{inv}},
\end{equation}
where $v_\theta$ is a pretrained conditional flow model, $x_t^{\mathrm{inv}}$ and $x_t^{\mathrm{gen}}$ are latent trajectories.

\textbf{Impulsive controlled generation.}~ Since flow models are not trained to jointly condition on source images and edit instructions, the generated outputs can deviate substantially from the source, resulting in poor fidelity. This limitation motivates the need for fine-grained control during generation. Leveraging the inherently discretized nature of flow-based sampling, we introduce step-wise control over the latent trajectory and formulate the editing process as an impulsively controlled dynamics.
\begin{equation}
\label{eq:impulsive_ode}
d x_t^{\mathrm{gen}} = v_\theta(x_t^{\mathrm{gen}}, t, c^{\mathrm{edit}})\,dt
+
\sum_{k=1}^{T} u_k\,\delta(t-t_k).
\end{equation}
Between two adjacent timesteps $(t_{k-1}, t_k)$, the trajectory follows the pretrained edit-conditioned dynamics. At discrete timesteps, we apply an impulsive control $u_k$ to the latent state. 

\textbf{Multi-objective control at each impulse.}~ Control updates generally affect both edit compliance and source fidelity. Since realistic image editing requires preserving instance-specific details, excessive edits can make the result indistinguishable from unconditional generation. We therefore formulate editing as a constrained, trajectory-level multi-objective optimization problem, enforcing a minimum fidelity requirement $\epsilon$ during generation. Let $r_{\text{edit}}(x,t)$ and $r_{\text{fid}}(x,t)$ be differentiable rewards for edit quality and fidelity. At each timestep $t_k$, we update the latent via
\begin{equation}
\tilde{x}_k = x_k^{\mathrm{gen}} + u_k,
\end{equation}
where $u_k$ is obtained by solving the following constrained local multi-objective control problem:
\begin{equation}
\label{eq:multi_obj_control}
u_k^\star \in \mathrm{Pareto}\left\{
\left(
r_{\mathrm{edit}}(x_k^{\mathrm{gen}}+u_k,t_k),
r_{\mathrm{fid}}(x_k^{\mathrm{gen}}+u_k,t_k)
\right)
:
r_{\mathrm{fid}}(x_k^{\mathrm{gen}}+u_k,t_k)\ge \epsilon
\right\},
\end{equation}
where the definition of Pareto set is deferred to Sec.~\ref{sec:def_pareto}.
Here, the fidelity constraint defines the feasible region, while the Pareto objective characterizes the local edit--fidelity trade-off within that region. Directly solving Eq.~(\ref{eq:multi_obj_control}) at every sampling step is impractical. We therefore approximate this local multi-objective control problem using tractable surrogate update rules, which are introduced in Sec.~\ref{sec:method}.

\section{Methodology}\label{sec:method}

Building on the preceding problem formulation, we detail its practical implementation by defining (a) the inversion, inversion-free initialization, and generation procedures, and (b) the differentiable rewards for editing compliance and source fidelity.

\textbf{Inversion, inversion-free, and generation trajectories.}
Let $\{t_k\}_{k=0}^{T}$ be a discretization with $t_0 < \cdots < t_T$.
We denote by $x_k^{\mathrm{inv}}$ the inversion state at time $t_k$. A forward Euler discretization of~\eqref{eq:inv_ode} gives
\begin{equation}
\label{eq:inv_step}
x_{k+1}^{\mathrm{inv}}
=
x_k^{\mathrm{inv}} + (t_{k+1}-t_k)\,v_\theta\!\left(x_k^{\mathrm{inv}}, t_k, c^{\mathrm{src}}\right),
\quad x_0^{\mathrm{inv}} = x^{\mathrm{src}}.
\end{equation}
Alternatively, an inversion-free initialization constructs the noisy state as
\begin{equation}
x_k^{\mathrm{noisy}} = (1 - t_k)\,x^{\mathrm{src}} + t_k\, N_k,
\quad N_k \sim \mathcal{N}(0, I).
\end{equation}
For editing, we initialize $x_T^{\mathrm{gen}} = x_T^{\mathrm{inv}}$ or $x_T^{\mathrm{gen}} = x_T^{\mathrm{noisy}}$, depending on the initialization strategy. The reverse-time explicit Euler discretization of Eq.~(\ref{eq:gen_ode}) is:
\begin{equation}
x_{k-1}^{\mathrm{gen}} 
=
x_k^{\mathrm{gen}} + (t_{k-1}-t_k)\,v_\theta\!\left(x_k^{\mathrm{gen}}, t_k, c^{\mathrm{edit}}\right),
\end{equation}
yielding $x_0^{\mathrm{gen}}$ as the edited image $x^{\mathrm{edit}}$. At each timestep $t_{k-1}$, we treat the current latent $x_{k-1}^{\mathrm{gen}}$ as the optimization variable and apply differentiable guidance updates for desired control.

\textbf{Reward definition via clean-state approximation.}~ 
Fidelity and edit compliance are defined on the underlying clean content, which is not directly accessible during intermediate steps. We therefore approximate the clean latent at each timestep using a single-step denoising estimate:
\begin{equation}
\hat{x}_0(x, t, c) := x - t\,v_\theta(x,t,c).
\end{equation}
Although approximate, this estimate is sufficient for computing informative gradients, as most reward functions operate on clean images.

We define \textbf{edit reward} as a similarity between the predicted clean content and the edit condition:
\begin{equation}
r_{\text{edit}}(x,t) := \mathrm{Sim}\big(\hat{x}_0(x,t, c^{\mathrm{edit}}),\, c^{\mathrm{edit}}\big).
\end{equation}
In practice, $\mathrm{Sim}(\cdot,\cdot)$ can be instantiated using different CLIP-based objectives, including text-image similarity, directional similarity, or contrastive differences, as summarized in Appendix~\ref{sec:edit_reward}.

We measure \textbf{fidelity reward} by the reconstruction error between the predicted clean content under the source condition and the original source latent:\begin{equation}r_{\text{fid}}(x,t) := - \big\| \hat{x}_0(x,t, c^{\mathrm{src}}) - x^{\mathrm{src}} \big\|_2^2.\end{equation}This encourages the generated trajectory to remain close to the source content while allowing controlled deviations for editing.

\remark{Fidelity is a proxy, not the target.}
Importantly, $r_{\text{fid}}$ is only an imperfect surrogate for perceptual or semantic fidelity. Using the original image as the reference is not fully justified when editable and preservable content are not explicitly separated: perfect fidelity would imply no edit at all. Thus, treating fidelity as a symmetric objective conflates the desired constraint of preserving what should remain unchanged with an imperfect reward proxy.

\textbf{Balancing edit alignment and source fidelity.}~ 
Edit alignment and source fidelity are inherently conflicting objectives: perfect fidelity leads to trivial reconstruction with no semantic modification. We therefore formulate image editing as a constrained multi-objective optimization problem over the generative trajectory, where edit rewards are maximized while preserving source consistency through explicit fidelity constraints. A direct linear scalarization 
$
\max_x\;
r_{\text{edit}}(x,t)
+
\lambda\, r_{\text{fid}}(x,t)$
is often brittle in practice, since the optimal trade-off parameter \(\lambda\) varies significantly across timesteps due to the changing geometry of the denoising trajectory. Moreover, edit and fidelity gradients may conflict locally, resulting in unstable or myopic updates. 
A simple constrained alternative first applies an \textbf{edit-first} update $x \gets x + \gamma\, \nabla_x r_{\text{edit}}(x,t)$,
which is accepted as long as the fidelity constraint
\(
r_{\text{fid}}(x_k^{\mathrm{gen}}, t_k) \ge \varepsilon
\)
remains satisfied. If the constraint is violated, approximate feasibility is pursued through projected correction steps,
$x \gets x + \eta\, \nabla_x r_{\text{fid}}(x,t)$,
repeated for at most \(J\) iterations. The finite correction budget bounds computation but does not guarantee exact feasibility at every inner iteration. This constrained refinement acts as a local trust-region mechanism, preventing excessive deviation from the source image while still allowing flexible semantic edits.

\begin{figure}[H]
  \centering
  \includegraphics[width=\linewidth]{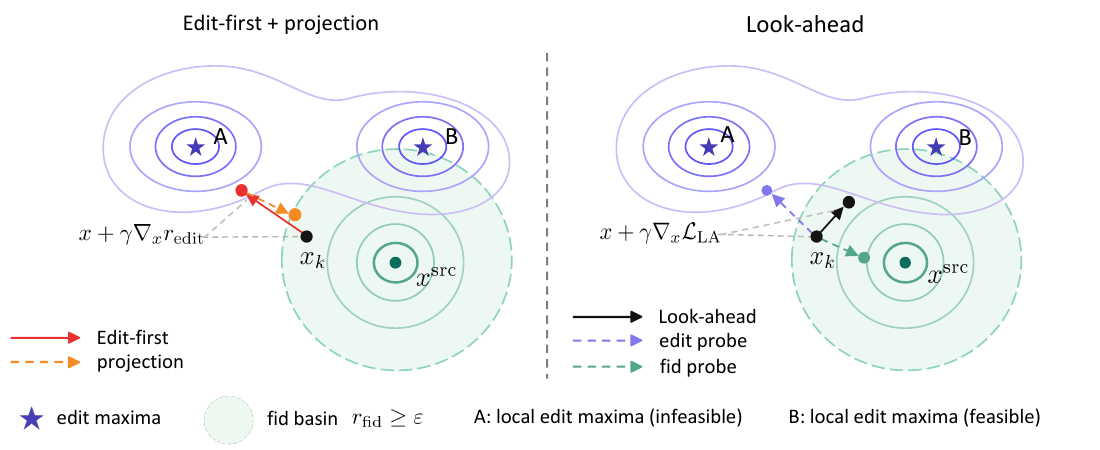}
  \caption{
    \textbf{Mechanism comparison.}
    \emph{Left}: edit-first update moves along $g_{\text{edit}}$ but may violate fidelity constrain and need projects back to the fidelity basin.
    \emph{Right}: the look-ahead update uses edit and fidelity probes to evaluate $\nabla_x \mathcal{L}_{\text{LA}}$ before committing an update, steering the trajectory toward the feasible edit maximum inside the fidelity basin.
  }
  \label{fig:mechanism}
  \vspace{-0.3cm}
\end{figure}

This strategy remains purely local and greedy, as each update is determined only by the current state. In diffusion and flow trajectories, such local updates can accumulate over time, causing semantic drift or source-fidelity degradation in later steps. Analogously, existing multi-objective optimization (MOO) methods, such as MGDA \cite{desideriMultiplegradientDescentAlgorithm2012} and PCGrad \citep{yuGradientSurgeryMultiTask2020}, are also designed to resolve gradient conflicts at the current optimization state. However, this local view is insufficient for flow-based image editing, where each update is
propagated through the remaining generation trajectory before affecting the final
image. As a result, a locally conflict-free direction may still induce an
undesirable future trajectory. 

\textbf{Look-ahead guidance.}~ To address this issue, we approximate the constrained MOO update with a look-ahead strategy that anticipates the effects of candidate updates on the predicted clean endpoints, followed by a bounded fidelity correction when necessary. Specifically, we first construct two look-ahead states
\begin{equation}
x_{\mathrm{edit}}(x)=x+\alpha\nabla_x r_{\mathrm{edit}}(x,t),\qquad
x_{\mathrm{fid}}(x)=x+\beta\nabla_x r_{\mathrm{fid}}(x,t),
\end{equation}
which represent edit-oriented and fidelity-oriented updates, respectively. We then define the look-ahead objective
\begin{equation}
\mathcal{L}_{\mathrm{LA}}(x):=
r_{\mathrm{edit}}(x_{\mathrm{edit}}(x),t)
+\lambda r_{\mathrm{fid}}(x_{\mathrm{fid}}(x),t),\label{eq:lookahead_objective}
\end{equation}
and update the latent using
\begin{equation}
x\leftarrow x+\gamma\nabla_x\mathcal{L}_{\mathrm{LA}}(x).
\end{equation}
Importantly, each reward in Eq. (13) is evaluated on the condition-specific predicted clean endpoint recomputed from the corresponding perturbed latent. Unlike direct scalarization, which combines reward gradients at the current latent, our look-ahead objective accounts for how estimated updates alter the predicted downstream rewards before committing to an actual update. This incorporates local curvature information arising from both the reward functions and the denoising map, yielding a locally non-myopic update without explicitly unrolling additional denoising steps. We provide further analysis in Appendix~\ref{app:additional_analyses}.

\paragraph{Full generation-phase algorithm.}
Starting from the initial latent, each generation timestep first applies an
edit-conditioned Euler step and then performs $p$ inner-loop trust-region
refinements. At each refinement step, either the edit-first or look-ahead update
rule proposes a latent update by evaluating rewards on the predicted clean state
$\hat{x}_0$ and its decoded image. The fidelity constraint then corrects updates
that induce excessive perceptual drift from the source appearance. The complete
procedure is given in Algorithm~\ref{alg:tr_edit}.

\section{Experiments}\label{sec:experiments}
We structure the experiments around four research questions. \textbf{RQ1}: How effective is our method for real-image editing versus baselines? \textbf{RQ2}: Does gradient conflict between edit and fidelity objectives arise during generation, and can look-ahead mitigate it? \textbf{RQ3}: Does our method align with its intended reward-space formulation? \textbf{RQ4}: How does the fidelity threshold $\varepsilon$ affect the edit--fidelity trade-off? We address them via main comparisons, gradient-conflict analysis, reward-space trajectory analysis, and $\varepsilon$-sensitivity analysis.

\subsection{Experimental Setup}\label{sec:setup}

We evaluate our algorithm and baselines on PIE-Bench v1~\cite{juDirectInversionBoosting2023}, a real-image editing benchmark spanning $10$ edit categories, including object replacement, addition, deletion, attribute editing, background modification, and style transfer. For fair comparison, we report results on this public benchmark, where all baselines generate valid outputs under a unified FLUX-based evaluation pipeline. When applicable, all methods use FLUX.1-dev~\cite{labs2025flux} as the backbone.

We compare against eight baselines. DDIM Inversion~\cite{songDenoisingDiffusionImplicit2020} inverts the source image to a noise latent via ODE and regenerates it with the edit prompt. TABA~\cite{staniszewskiThereBackAgain2026} progressively edits the image by alternating forward noising and edit-prompt denoising. RF-Inversion~\cite{routSemanticImageInversion2024} uses a stochastic ODE for inversion and injects source features during generation to preserve structure. RF-Solver-Edit~\cite{wangTamingRectifiedFlow2024} maintains fidelity by injecting V-values from the source branch into the edit branch. FlowEdit~\cite{kulikovFlowEditInversionFreeTextBased2025} is inversion-free and shifts the trajectory using the velocity difference between source and edit conditions. FlowChef-Edit~\cite{patelSteeringRectifiedFlow2024} steers the rectified-flow trajectory with gradient guidance and attention-based region masks. FlowAlign~\cite{kimFlowAlignTrajectoryRegularizedInversionFree2025} regularizes the editing trajectory toward the source flow. ITOC~\cite{changTrainingFreeRewardGuidedImage2026} inverts the source image and optimizes the full trajectory control with a terminal reward via adjoint states. We adapt the official implementations of FlowAlign and ITOC to the same FLUX backbone. DDIM Inversion and TABA are also adapted to flow models in our evaluation. When available, all baselines use the default hyperparameters from their official repositories. For fairness, we report the ConceptAttention (CA) variant of FlowChef-Edit, which uses masks automatically extracted by ConceptAttention rather than human annotations.

For our method, we use inversion-free initialization, constructing the initial noisy state by forward flow without separate inversion. We run FLUX.1-dev for $28$ generation steps and apply multi-objective test-time optimization at every denoising step. Each latent is refined with three inner optimization iterations before advancing to the next timestep. The edit objective is a CLIP-contrastive reward. During generation, edit and fidelity gradients are computed on a single-step clean-latent prediction $x_0$ from the current noisy latent. Source preservation uses a latent-$\ell_2$ fidelity measure with threshold $\varepsilon=-0.1$. When the fidelity constraint is violated, we apply correction updates with step size $\eta=5.0$, for at most three correction steps per denoising step. We use $\ell_2$-normalized gradients and set the edit probe, fidelity probe, and committed-update step sizes to $\alpha=\beta=\gamma=5.0$, with fidelity weight $\lambda=1$. Full sensitivity analyses are provided in Appendix~\ref{app:sensitivity_ablations}.

Following prior work~\cite{juDirectInversionBoosting2023,patelSteeringRectifiedFlow2024}, we evaluate source preservation on unedited regions using structure distance, PSNR, LPIPS, MSE, and SSIM. These metrics capture complementary fidelity aspects: structure distance measures spatial-layout preservation; PSNR and MSE assess pixel-level reconstruction; LPIPS measures learned perceptual similarity; and SSIM evaluates luminance, contrast, and structural consistency. Better preservation corresponds to higher PSNR and SSIM, and lower structure distance, LPIPS, and MSE. For edit quality, we report CLIP similarity between the edited image and target prompt, computed on both the whole image and edited region. Whole-image CLIP measures global text-image alignment, while edited-region CLIP assesses whether the intended modification is realized in the target area.

\subsection{Quantitative Comparison}
Table~\ref{tab:main_results} summarizes the quantitative comparison. Our method achieves the best structure distance and LPIPS on unedited regions, as well as the best whole-image target-prompt alignment. ITOC~\cite{changTrainingFreeRewardGuidedImage2026} attains the best PSNR, MSE, and SSIM, but at substantially lower edited-region alignment, and our method ranks second on PSNR and MSE. For edited-region CLIP similarity, it ranks third behind TABA~\cite{staniszewskiThereBackAgain2026} and DDIM Inversion~\cite{songDenoisingDiffusionImplicit2020}. However, these methods lie at the high-edit extreme of the trade-off, improving local alignment at the cost of much worse preservation. In contrast, our method reaches a more favorable frontier point, strongly preserving unedited content while surpassing all baselines in global edit alignment.

\begin{table*}[t]
\caption{Quantitative comparison on the PIE-Bench~\cite{juDirectInversionBoosting2023}. Our method achieves competitive preservation metrics while also obtaining the best global target-prompt alignment. Best results are in \textbf{bold} and second-best results are \underline{underlined}.}
\label{tab:main_results}
\centering
\small
\resizebox{\textwidth}{!}{
\begin{tabular}{lccccccc}
\toprule
\multirow{2}{*}{Method} & \multirow{2}{*}{Structure $\downarrow$} & \multicolumn{4}{c}{Background Preservation} & \multicolumn{2}{c}{CLIP Similarity} \\
\cmidrule(lr){3-6} \cmidrule(lr){7-8}
 &  & PSNR $\uparrow$ & LPIPS $\downarrow$ & MSE $\downarrow$ & SSIM $\uparrow$ & Whole $\uparrow$ & Edited $\uparrow$ \\
\midrule
DDIM Inversion~\cite{songDenoisingDiffusionImplicit2020} & 0.1229 & 12.7867 & 0.3578 & 0.0763 & 0.6025 & 26.7718 & \underline{23.9214} \\
TABA~\cite{staniszewskiThereBackAgain2026} & 0.1233 & 12.8892 & 0.3496 & 0.0744 & 0.6056 & 27.4278 & \textbf{24.2488} \\
RF-Inversion~\cite{routSemanticImageInversion2024} & 0.0561 & 19.1570 & 0.2290 & 0.0171 & 0.6600 & 26.1720 & 23.3210 \\
RF-Solver-Edit~\cite{wangTamingRectifiedFlow2024} & 0.0249 & 24.4093 & 0.1149 & 0.0057 & \underline{0.8446} & 25.6668 & 22.7751 \\
FlowChef-Edit~\cite{patelSteeringRectifiedFlow2024} & 0.0395 & 24.3100 & 0.1239 & -- & 0.7960 & 25.3200 & 22.3400 \\
FlowEdit~\cite{kulikovFlowEditInversionFreeTextBased2025} & 0.0276 & 22.0007 & \underline{0.1068} & 0.0094 & 0.8426 & 26.0726 & 23.1487 \\
FlowAlign~\cite{kimFlowAlignTrajectoryRegularizedInversionFree2025} & 0.0979 & 14.6995 & 0.2599 & 0.0469 & 0.6977 & 27.0154 & 23.2178 \\
ITOC~\cite{changTrainingFreeRewardGuidedImage2026} & \underline{0.0188} & \textbf{26.5574} & 0.1125 & \textbf{0.0042} & \textbf{0.8513} & \underline{27.6348} & 21.6832 \\
Ours                     & \textbf{0.0176} & \underline{25.0050} & \textbf{0.0957} & \underline{0.0045} & 0.8433 & \textbf{28.6262} & 23.7916 \\
\bottomrule
\end{tabular}
}
\end{table*}

\subsection{Qualitative Comparison}

\begin{figure*}[t]
  \centering
  \includegraphics[width=\textwidth]{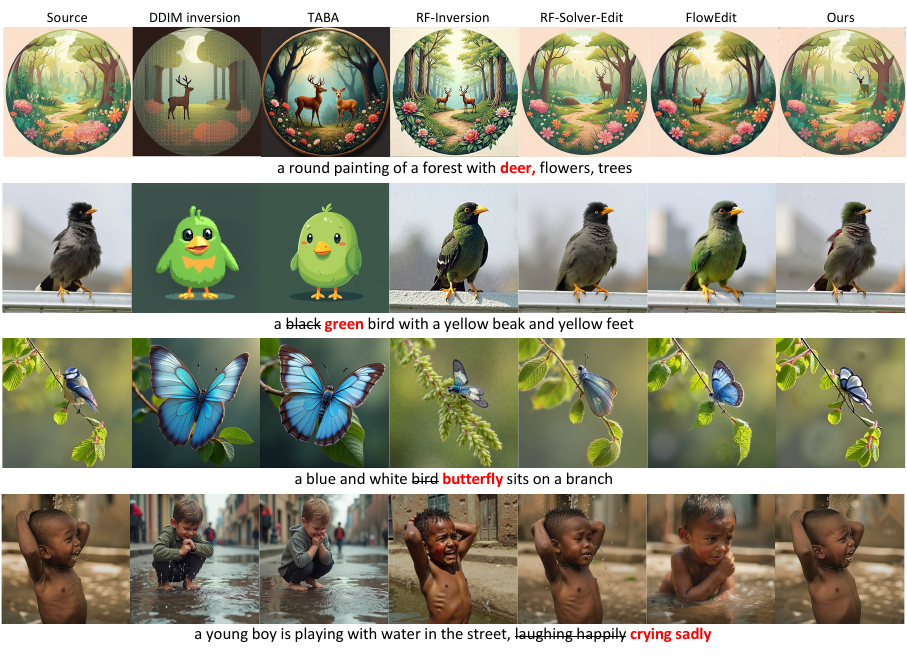}
  \caption{
    \textbf{Qualitative comparison on representative PIE-Bench edit categories.}
    Our method consistently completes the requested edit while preserving unedited content.
  }
  \label{fig:qualitative}
  \vspace{-0.3cm}
\end{figure*}

As shown in Figure~\ref{fig:qualitative}, our method produces edits that are more faithful to the target instructions while better preserving the source structure and visual realism than competing methods. 
In the deer-addition example, our method inserts the deer at an appropriate location while leaving the surrounding trees, flowers, and road almost unchanged. Competing methods, by contrast, nearly regenerate a new forest background. For the object-attribute editing task, our method changes the bird's feathers to green precisely. FlowEdit~\cite{kulikovFlowEditInversionFreeTextBased2025} also applies the color change, but the bird's shape is altered substantially. Similarly, in the object-replacement example, baselines such as RF-Solver-Edit~\cite{wangTamingRectifiedFlow2024} and RF-Inversion~\cite{routSemanticImageInversion2024} fail to preserve the background leaves, whereas our method generates a butterfly with the correct color while keeping the leaves unchanged. In the facial-expression editing example, most competing methods almost replace the child entirely. Our method instead achieves the semantic edit through subtle changes around the eyes and face, without the background corruption and identity drift observed in other methods.

\subsection{Diagnostic Analysis}

\subsubsection{Gradient Conflict along the Generation Trajectory}

\begin{figure*}[h]
  \centering
  \includegraphics[width=\textwidth]{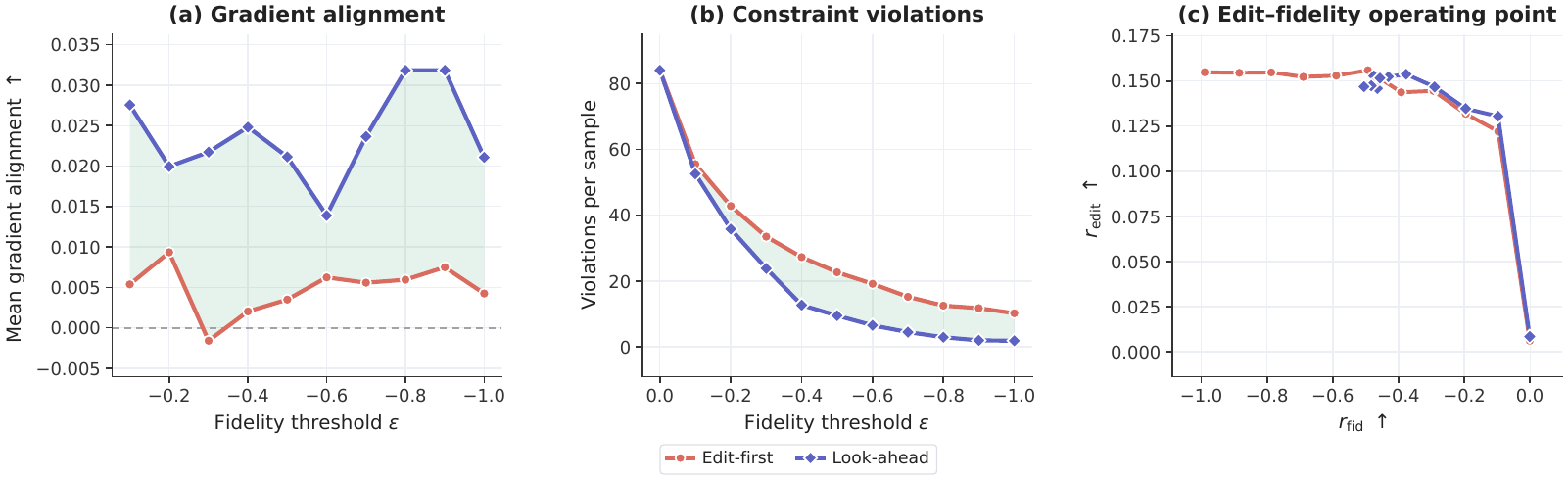}
  \caption{
    \textbf{Gradient conflict and look-ahead analysis} across PIE-Bench samples.
    \emph{(a)} Mean per-step alignment $\frac{1}{T}\sum_k c_k$ vs.\ $\varepsilon$: the look-ahead proposal achieves consistently higher alignment than the edit proposal.
    \emph{(b)} Violations per sample vs.\ $\varepsilon$: the look-ahead rule triggers substantially fewer fidelity violations, especially at loose thresholds.
    \emph{(c)} $(r_{\text{fid}},\, r_{\text{edit}})$ operating points: both rules trace nearly the same curve, confirming that fewer violations do not degrade edit quality.
  }
  \label{fig:grad_conflict}
\end{figure*}

We analyze the gradient conflict between edit and fidelity objectives along the generation trajectory, and show that the proposed look-ahead mechanism improves gradient alignment.
At generation step $k$, we measure gradient alignment by the cosine similarity $c_k = \mathrm{cos}\!\left(\nabla_x r_{\text{edit}}(x_k,t_k),\, \nabla_x r_{\text{fid}}(x_k,t_k)\right)$, where $x_k$ is the latent at timestep $t_k$. Figure~\ref{fig:grad_conflict}(a) reports the mean per-step gradient alignment $\frac{1}{T}\sum_k c_k$ as a function of $\varepsilon$. The look-ahead proposal consistently achieves higher alignment than the edit proposal across all fidelity thresholds.

When the edit gradient and fidelity gradient are better aligned, the proposed update is less likely to cross the fidelity threshold $\varepsilon$.
Figure~\ref{fig:grad_conflict}(b) confirms this: the look-ahead proposal incurs substantially fewer violations per sample than the edit proposal across all $\varepsilon$ values, particularly at loose thresholds.

Figure~\ref{fig:grad_conflict}(c) shows that the $(r_{\text{fid}},\, r_{\text{edit}})$ operating points of the two rules trace nearly the same curve across the $\varepsilon$ sweep.
This says that the proposed look-ahead mechanism reduces the number of fidelity violation without sacrificing edit quality.
We also observe that the look-ahead operating points saturate around $r_{\text{fid}} \approx -0.5$ regardless of how loose $\varepsilon$ becomes, unlike edit-first which continues to degrade fidelity.
This is because the look-ahead gradient inherently couples both objectives at every step, preventing the update direction from becoming purely edit-driven even under a slack constraint.

\begin{figure*}[h]
  \centering
  \includegraphics[width=\textwidth]{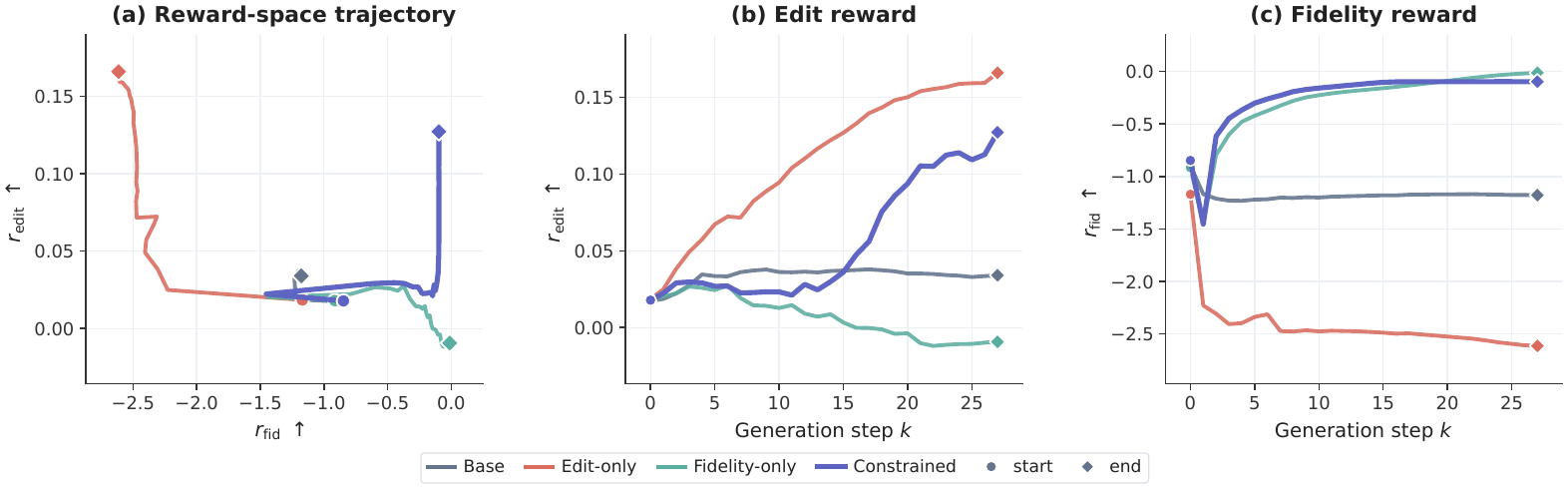}
  \caption{
    \textbf{Reward-space trajectories under different generation-time update rules.}
    \emph{(a)} Trajectories in the $(r_{\text{fid}}, r_{\text{edit}})$ plane; \emph{(b)} edit reward; \emph{(c)} fidelity reward over generation steps.
    Edit-only updates improve $r_{\text{edit}}$ aggressively but severely degrade fidelity; fidelity-only updates preserve source consistency but suppress editing.
    Our method reaches a balanced trajectory, improving edit reward while keeping fidelity degradation controlled.
  }
  \label{fig:reward_by_mode}
  \vspace{-0.3cm}
\end{figure*}

\subsubsection{Reward-Space Trajectories under Different Update Rules}
Figure~\ref{fig:reward_by_mode} compares four generation-time update rules in reward space: the base trajectory without optimization, edit-only updates, fidelity-only updates, and our proposed method. The comparison makes the role of the proposed method explicit. Edit-only updates improve $r_{\text{edit}}$ most aggressively, but do so by driving the trajectory to a severely degraded fidelity regime. Fidelity-only updates move in the opposite direction: they recover source consistency, but suppress semantic editing and eventually reduce edit reward. The base trajectory rises slightly in the direction of editing reward, mainly because the generated trajectory is controlled by the target condition, but decreases in the dimension of fidelity reward.

Our method follows a more balanced path. As shown in Fig.~\ref{fig:reward_by_mode}(a), it moves the trajectory toward substantially higher edit reward without collapsing fidelity to the extremes reached by edit-only updates. Panels (b) and (c) make the same behavior visible over time: $r_{\text{edit}}$ increases steadily, while $r_{\text{fid}}$ initially drops as semantic changes emerge and then stabilizes at a controlled level rather than drifting arbitrarily. This behavior is consistent with our formulation. The method performs edit-first optimization under a fidelity threshold.

\subsubsection{The Edit--Fidelity Trade-Off under Different Fidelity Thresholds}

\begin{figure*}[h]
  \centering
  \includegraphics[width=\textwidth]{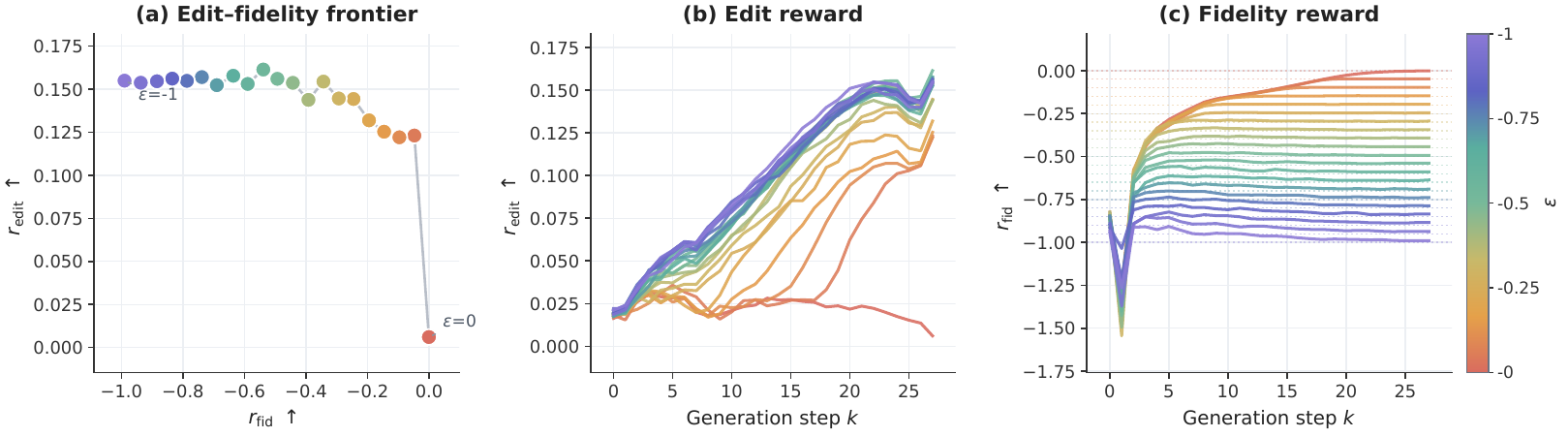}
  \caption{
    \textbf{Effect of the fidelity threshold $\varepsilon$ on the edit--fidelity trade-off.}
    \emph{(a)} $(r_{\text{fid}}, r_{\text{edit}})$ per $\varepsilon$: the connected curve forms an empirical curve parameterized directly by $\varepsilon$.
    \emph{(b,c)} Edit and fidelity reward over generation steps; dotted lines in \emph{(c)} mark the constraint boundary. 
  }
  \label{fig:eps_sweep}
  \vspace{-0.2cm}
\end{figure*}

Figure~\ref{fig:eps_sweep} illustrates how the fidelity threshold $\varepsilon$ shapes the reward-space trajectory across PIE-Bench samples under the edit-first proposal rule used in all runs.
The left panel traces the $(r_{\text{fid}}, r_{\text{edit}})$ operating point of each run as $\varepsilon$ varies.
The curve sweeps continuously from the upper-left (small $\varepsilon$, loose constraint, high edit reward) to the lower-right (large $\varepsilon$, strict constraint, high fidelity reward), forming an empirical curve parameterized directly by $\varepsilon$.

Two qualitative regimes are visible.
For $\varepsilon \leq -0.5$, the edit reward saturates: loosening the constraint further provides diminishing returns in edit quality while fidelity continues to degrade.
Above $\varepsilon \approx -0.25$, the edit reward decreases gradually as the fidelity projection increasingly overrides the edit gradient in the low-noise stage of generation.
At $\varepsilon = 0$ the constraint is so strict that the method reduces to approximate reconstruction, and edit reward collapses to near zero.

These observations do not prescribe a single optimal $\varepsilon$; the appropriate setting depends on the task.
For image editing, fidelity and edit compliance are both essential: a value that preserves unedited regions while still completing the requested semantic change.
We find that $\varepsilon = -0.1$ on the latent-$\ell_2$ fidelity metric achieves this balance in practice.
Panels~(b) and~(c) of Figure~\ref{fig:eps_sweep} show that this setting produces a steady increase in $r_{\text{edit}}$ through the generation trajectory while keeping $r_{\text{fid}}$ above the constraint boundary, which is consistent with the PIE-Bench preservation results reported in Table~\ref{tab:main_results}.

\begin{table}[h]
\centering
\small
\caption{Inference runtime and peak GPU memory.}
\label{tab:runtime_memory}
\label{tab:optimization_runtime}
\begin{tabular}{lcc}
\toprule
Method & Runtime / image (s) $\downarrow$ & Peak memory (GiB) $\downarrow$ \\
\midrule
FlowEdit~\cite{kulikovFlowEditInversionFreeTextBased2025} & \textbf{7.00} & 34.18 \\
TABA~\cite{staniszewskiThereBackAgain2026} & 7.80 & \textbf{32.05} \\
DDIM Inversion~\cite{songDenoisingDiffusionImplicit2020} & 8.08 & \textbf{32.05} \\
RF-Solver-Edit~\cite{wangTamingRectifiedFlow2024} & 16.45 & 32.59 \\
RF-Inversion~\cite{routSemanticImageInversion2024} & 25.90 & 35.72 \\
ITOC~\cite{changTrainingFreeRewardGuidedImage2026} & 139.25 & 40.93 \\
Ours & 185.88 & 55.82 \\
\bottomrule
\end{tabular}
\end{table}

\subsubsection{Computational Cost}
Table~\ref{tab:runtime_memory} reports per-image runtime and peak GPU memory on a single NVIDIA H200 GPU. Methods that perform reward-gradient optimization, including ITOC~\cite{changTrainingFreeRewardGuidedImage2026} and ours, incur higher costs than baselines with fixed inversion and generation procedures. Our additional overhead comes from evaluating both objectives and applying iterative fidelity corrections. As detailed in Appendix~\ref{app:computational_cost}, linearized projection reduces runtime to 139.4\,s. Improving efficiency remains important future work.

\begin{table}[H]
\centering
\small
\caption{Edit-first versus look-ahead under the same per-image projection budget.}
\label{tab:update_rule_matched_budget}
\resizebox{\textwidth}{!}{
\begin{tabular}{lccccccc}
\toprule
Update rule & Structure $\downarrow$ & PSNR $\uparrow$ & LPIPS $\downarrow$ & MSE $\downarrow$ & SSIM $\uparrow$ & CLIP Similarity (Whole) $\uparrow$ & CLIP Similarity (Edited) $\uparrow$ \\
\midrule
Edit-first & 0.0138 & 26.2152 & 0.0841 & 0.0033 & 0.8258 & 26.9845 & \textbf{22.0966} \\
Look-ahead & \textbf{0.0091} & \textbf{27.2950} & \textbf{0.0751} & \textbf{0.0026} & \textbf{0.8476} & \textbf{27.2370} & 22.0130 \\
\bottomrule
\end{tabular}}
\end{table}

\subsubsection{Ablation Studies}
We summarize key ablations below; Appendix~\ref{app:additional_analyses} reports full results and a user study.

\noindent\textbf{Update rule.} Table~\ref{tab:update_rule_full} shows that edit-first and look-ahead reach nearly identical endpoints on the full PIE-Bench under unrestricted correction, suggesting that the constrained framework accounts for most of the overall gain. Under a matched per-image projection budget, Table~\ref{tab:update_rule_matched_budget} shows that look-ahead improves every preservation metric, reaching a favorable edit--fidelity operating point with less reliance on corrective projections.

\noindent\textbf{Hyperparameters.} Increasing $p$ or $J$ beyond 3 adds cost for marginal gains (Table~\ref{tab:p_j_sensitivity}). Performance is stable over a $16\times$ range of $\lambda$ and probe distances $\alpha,\beta\in[1,10]$ (Table~\ref{tab:lookahead_parameter_sensitivity}). We use $p=J=3$ and $\lambda=1$ as practical defaults.

\noindent\textbf{Rewards and initialization.} ImageReward and HPS v2.1 yield a similar edit--fidelity balance to CLIP-contrastive guidance (Table~\ref{tab:guidance_reward_ablation}). Inversion-free initialization improves every preservation metric over inversion (Table~\ref{tab:reward_initialization_ablation}).

\section{Conclusion}
We have presented a principled, inversion-free framework for text-guided image editing that formulates the problem as constrained multi-objective optimization over the generative trajectory. The framework evaluates candidate interventions through their predicted clean endpoints and enforces an explicit source-fidelity constraint. The look-ahead update improves gradient alignment and reduces reliance on corrective projections, while the constraint exposes a controllable edit--fidelity operating point. Experiments demonstrate stronger source preservation and target alignment than existing inversion-based and inversion-free methods, with the trade-off of additional inference-time computation.

\begin{ack}
    Doudou Zhang and Qi Chen are supported by research
    funding from the ELLIS Institute Finland and the Department of Computer Science
    at Aalto University. We acknowledge CSC-IT Center for Science, Finland, for
    providing access to the supercomputers Roihu and LUMI, owned by the European
    High Performance Computing Joint Undertaking (EuroHPC JU) and hosted by CSC
    Finland in collaboration with the LUMI consortium. We also acknowledge the
    computational resources provided by the Aalto Science-IT project through the
    Triton cluster.
\end{ack}

\bibliographystyle{plainnat}

\bibliography{neurips_2026}

\newpage
\appendix

\section{Concepts in multi-objective optimization}\label{sec:def_pareto}

\begin{definition}(Pareto Dominance.)
For any two vectors $\vect{r}^1, \vect{r}^2 \in \Ocal$, we denote $\vect{r}^1 \preceq \vect{r}^2$ if $\vect{r}^2 - \vect{r}^1 \in \Reals_+^n$, which implies $r_i^1 \leq r_i^2, \forall i \in [n]$. If there further exists at least one $j\in[n]$ s.t. $r_j^1 < r_j^2$, we have $\vect{r}^1 \prec \vect{r}^2$.
Then, we say a solution $x_1 \in \Xcal$ dominates another solution $x_2\in \Xcal$ iff $\vect{r}(x_1) \prec \vect{r}(x_2)$. Note if $\vect{r}(x_1) \not\prec \vect{r}(x_2)$, $x_1$ does not dominate $x_2$.  
\end{definition}

\begin{definition}(Pareto Optimal.)
A solution $x^*\in\Xcal$ is Pareto optimal if it is not dominated by any other solution, \ie, $\vect{r}(x) \not\prec \vect{r}(x^*), \forall x \in \Xcal - \{x^*\}$.
\end{definition}

\begin{definition}(Pareto Set and Pareto Front.) The set of all the Pareto optimal solutions is given by $\Pcal \eqdef \mathrm{Pareto}\{ x^*: \forall x \in \Xcal - \{x^*\}, \vect{r}(x) \not\prec \vect{r}(x^*)\}$. Then, the corresponding set of multi-objective values of the Pareto set is called the Pareto front, denoted by $\Fcal \eqdef \{\vect{r}(x^*), \forall x^* \in \Pcal\}$.

\end{definition}
\section{Algorithm}

\begin{adjustbox}{width=\textwidth}
\begin{minipage}{0.8\textwidth}
\begin{algorithm}[H]
\caption{Constrained Look-Ahead Reward-guided Edit}
\label{alg:tr_edit}
\small
\begin{algorithmic}[1]
\REQUIRE Source latent $x^{\mathrm{src}}$, initial noisy latent $x_T^{\mathrm{gen}}$, source condition $c^{\mathrm{src}}$, edit condition $c^{\mathrm{edit}}$,
timesteps $\{t_k\}_{k=0}^{T}$, inner steps $p$, edit step size $\gamma$,
fidelity threshold $\varepsilon$, projection step size $\eta$, max projection steps $J$
\ENSURE Edited image $x^{\mathrm{edit}} = x_0^{\mathrm{gen}}$
\FOR{$k = T,T-1,\ldots,1$}
  \STATE $x \gets x_k^{\mathrm{gen}} + (t_{k-1}-t_k)\,v_\theta(x_k^{\mathrm{gen}},\,t_k,\,c^{\mathrm{edit}}), t \gets t_{k-1}$
  \FOR{$i = 0,1,\ldots,p-1$}
    \STATE \textbf{// \underline{Version 1} (Edit-first):}
    \STATE \colorbox{blue!15}{$x_{\mathrm{trial}} \gets x + \gamma \nabla_x r_{\text{edit}}(x,t)$}
    \STATE \textbf{// \underline{Version 2} (Look-ahead):}
    \STATE \colorbox{violet!15}{$x_{\mathrm{edit}}(x) \gets x + \alpha\,\nabla_x r_{\text{edit}}(x,t)$}
    \STATE \colorbox{violet!15}{$x_{\mathrm{fid}}(x) \gets x + \beta\,\nabla_x r_{\text{fid}}(x,t)$}
    \STATE \colorbox{violet!15}{$\mathcal{L}_{\mathrm{LA}}(x) \gets r_{\text{edit}}(x_{\mathrm{edit}}(x), t) + \lambda\, r_{\text{fid}}(x_{\mathrm{fid}}(x), t))$}
    \STATE \colorbox{violet!15}{$x_{\mathrm{trial}} \gets x + \gamma\,\nabla_x \mathcal{L}_{\mathrm{LA}}(x)$}
    \IF{$r_{\text{fid}}(x_{\mathrm{trial}},t) \geq \varepsilon$}
      \STATE $x \gets x_{\mathrm{trial}}$
    \ELSE
      \STATE $x \gets x_{\mathrm{trial}}$
      \FOR{$j = 1,\ldots,J$}
        \IF{$r_{\text{fid}}(x,t) \geq \varepsilon$}
          \STATE \textbf{break}
        \ENDIF
        \STATE $x \gets x + \eta\,\nabla_x r_{\text{fid}}(x,t)$
      \ENDFOR
    \ENDIF
  \ENDFOR
  \STATE $x_{k-1}^{\mathrm{gen}} \gets x$
\ENDFOR
\RETURN $x^{\mathrm{edit}} \gets x_0^{\mathrm{gen}}$
\end{algorithmic}
\end{algorithm}
\end{minipage}
\end{adjustbox}

\newpage

\section{Edit reward options}
\label{sec:edit_reward}

We instantiate the edit reward $r_{\text{edit}}(x,t)$ using CLIP-based
similarity scores computed on the predicted clean image:
\[
\hat{x}_0(x,t,c^{\mathrm{edit}}).
\]
Let $\phi_{\mathrm{img}}$ denotes the
CLIP image encoder, and $\phi_{\mathrm{txt}}$ denotes the CLIP text encoder.
We use cosine similarity
\[
\mathrm{cos}(a,b) := \frac{\langle a,b\rangle}{\|a\|_2\|b\|_2}.
\]
The edit reward can then be instantiated in several ways.

\paragraph{Text-image similarity.}
The simplest option directly measures the CLIP similarity between the predicted
clean image and the target edit prompt:
\begin{equation}
r_{\mathrm{clip\_text}}
=
\mathrm{cos}\!\left(
\phi_{\mathrm{img}}(\hat{x}_0),
\phi_{\mathrm{txt}}(c^{\mathrm{edit}})
\right).
\end{equation}
This reward encourages the generated image to align with the target edit
description.

\paragraph{Directional similarity.}
Following the common CLIP-directional formulation, we can instead compare the
image-space change with the corresponding text-space change:
\begin{equation}
r_{\mathrm{clip\_directional}}
=
\mathrm{cos}\!\left(
\phi_{\mathrm{img}}(\hat{x}_0) - \phi_{\mathrm{img}}(x^{\mathrm{src}}),
\phi_{\mathrm{txt}}(c^{\mathrm{edit}}) - \phi_{\mathrm{txt}}(c^{\mathrm{src}})
\right),
\end{equation}
where $x^{\mathrm{src}}$ is the source image. This reward encourages the edit
direction in image-embedding space to match the semantic change specified by
the source and target prompts.

\paragraph{Contrastive similarity.}
We also consider a contrastive variant that increases similarity to the target
prompt while decreasing similarity to the source prompt:
\begin{equation}
r_{\mathrm{clip\_contrastive}}
=
\mathrm{cos}\!\left(
\phi_{\mathrm{img}}(\hat{x}_0),
\phi_{\mathrm{txt}}(c^{\mathrm{edit}})
\right)
-
\mathrm{cos}\!\left(
\phi_{\mathrm{img}}(\hat{x}_0),
\phi_{\mathrm{txt}}(c^{\mathrm{src}})
\right).
\end{equation}
This reward is useful when direct target-prompt similarity alone is insufficient
to encourage a clear semantic departure from the source prompt.

\section{Experimental Details}
\label{app:experimental_details}

All experiments use FLUX.1-dev~\cite{labs2025flux} as the base generative model.
We run 28 denoising steps with guidance scale $3.5$ for target condition.
For each denoising step we apply $p=3$ inner optimization iterations with edit step size $\gamma=5.0$.
The edit and fidelity probe distances are $\alpha=\beta=5.0$, and all proposal gradients are $\ell_2$ normalized before applying their respective step sizes.
The fidelity projection uses step size $\eta=5.0$ and a maximum of $J=3$ correction steps per inner iteration.
For Look-ahead mechanism, fidelity weight uses $\lambda_\text{{fid}}=1$.
The fidelity threshold is fixed at $\varepsilon=-0.1$ (latent-$\ell_2$ metric) for main experiments.
The edit reward is CLIP-contrastive.
Forward initialization uses the noise level $\sigma_T$ corresponding to the first denoising step of FLUX.1-dev.

\section{Additional Results on the FlowEdit Evaluation Set}
\label{app:flowedit_quant}

FlowEdit~\cite{kulikovFlowEditInversionFreeTextBased2025} evaluates text-based real-image editing on a diverse collection of real images paired with source prompts and handcrafted target prompts. We follow this setting and report additional quantitative and qualitative results on all 273 examples of the released FlowEdit evaluation set. On this set, we use fidelity threshold $\varepsilon=-0.15$ and keep all other hyperparameters identical to the PIE-Bench experiments.
We report CLIP-T for target-text alignment and LPIPS, DINO, and DreamSim for source preservation.
We omit CLIP-I (source-to-edited image cosine similarity in CLIP space) from the table: as shown by~\cite{kuVIEScoreExplainableMetrics2024}, CLIP-I correlates poorly with human judgments of editing quality because CLIP embeddings operate at a coarse semantic level and are insensitive to pixel-level reconstruction errors, causing methods with poor fidelity to score spuriously high.
Since this evaluation set does not provide edit masks, we do not report mask-based edit-region metrics.

\begin{table*}[h]
\centering
\small
\caption{Quantitative comparison on the FlowEdit evaluation set. LPIPS, DINO, and DreamSim measure source preservation, while CLIP-T measures target-text alignment. Best results are in \textbf{bold}.}
\label{tab:flowedit_metrics}
\begin{tabular}{lcccc}
\toprule
Method & LPIPS $\downarrow$ & DINO $\uparrow$ & DreamSim $\downarrow$ & CLIP-T $\uparrow$ \\
\midrule
FlowEdit~\cite{kulikovFlowEditInversionFreeTextBased2025} & 0.1854 & 0.7402 & 0.2576 & 29.2862 \\
Ours & \textbf{0.1573} & \textbf{0.8443} & \textbf{0.1144} & \textbf{29.6528} \\
\bottomrule
\end{tabular}
\end{table*}

\begin{figure*}[h]
  \centering
  \includegraphics[width=\textwidth]{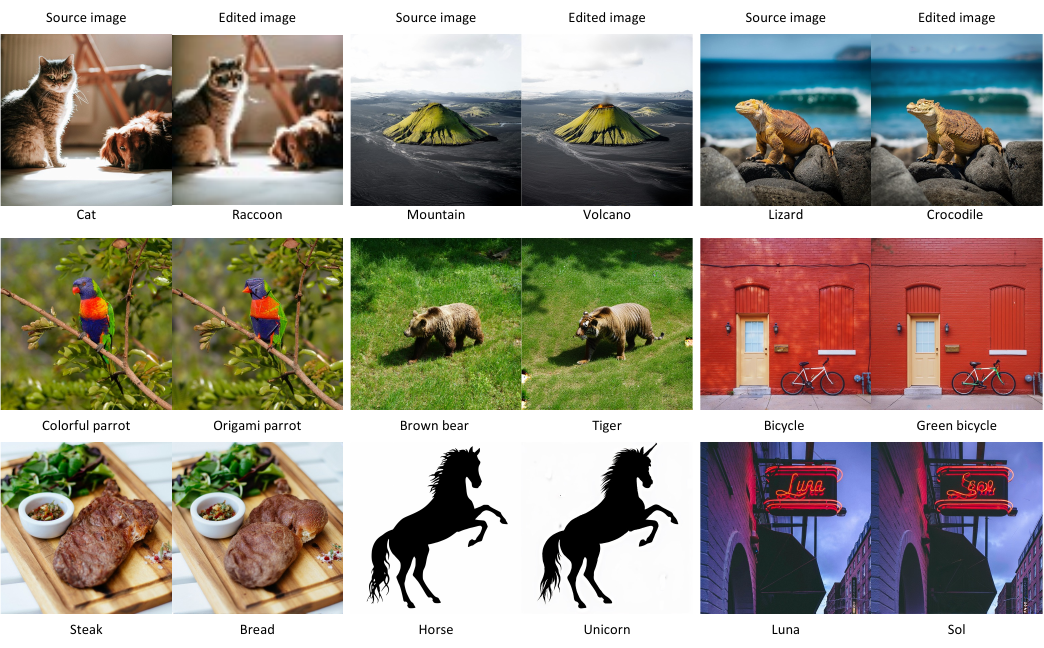}
  \caption{
    \textbf{Additional qualitative results on the FlowEdit evaluation set.} Our method executes diverse edit types, including object replacement, attribute modification, and style transfer, while preserving the structure of the original images.
  }
  \label{fig:flowedit_qual}
\end{figure*}

Our method achieves higher target-text alignment and better source preservation than FlowEdit according to CLIP-T, LPIPS, DINO, and DreamSim.
Figure~\ref{fig:flowedit_qual} provides qualitative examples on the same evaluation set, spanning object replacement, color and attribute editing, and style transfer.

\section{Additional Analyses}
\label{app:additional_analyses}

\paragraph{Scope of the trajectory objective.}
The Pareto formulation in Eq.~(5) defines a solution set rather than a directly computable update. Both rewards are nonlinear functions of a high-dimensional latent, the pretrained flow model, the clean-state prediction, and, for image-space rewards, the decoder and reward network. Approximating the full Pareto set independently at every sampling step would require repeated optimization under many scalarization weights or constraints. Our $\varepsilon$-constraint selects a desired operating region, while $p$ inner updates and at most $J$ corrections provide a tractable local approximation rather than an exact solution of the full Pareto set.

Our use of a predicted clean state is a compatibility choice that permits training-free use of existing differentiable image-space rewards. The framework does not require this representation: a differentiable timestep-aware reward $r_t(x_t)$ can replace $R(\hat{x}_0(x_t,t))$ without changing the constrained proposal and correction structure. Such rewards may provide denser intermediate feedback, but require reward models calibrated to the latent space and noise schedule.

\paragraph{Post-intervention clean-endpoint prediction.}
The rewards in Eq.~\eqref{eq:lookahead_objective} are not evaluated directly on the probe latents. For objective $i\in\{\mathrm{edit},\mathrm{fid}\}$, let
\begin{equation}
\phi_i(x):=\hat{x}_0(x,t,c_i),
\qquad
r_i(x,t):=R_i\!\left(\phi_i(x)\right),
\qquad
g_i(x):=\nabla_x r_i(x,t),
\end{equation}
where $c_i$ is the condition associated with the objective and $R_i$ includes the differentiable image-space reward computation. Given probe distance $s_i\in\{\alpha,\beta\}$, we form
\begin{equation}
x_i=x+s_i g_i(x)
\end{equation}
and recompute $\phi_i(x_i)$ before evaluating the post-intervention reward. The probe therefore asks how a hypothetical intervention at the current timestep changes the clean endpoint predicted from the remaining generation dynamics. This differs from current-state scalarization and gradient-combination rules, which use objective gradients evaluated only at the unperturbed latent.

Exact differentiation through a probe gives
\begin{equation}
\nabla_x r_i\!\left(x+s_i g_i(x),t\right)
=
\left(I+s_i H_i(x)\right)^{\!\top} g_i(x_i),
\qquad
H_i(x):=\nabla_x^2 r_i(x,t),
\label{eq:exact_probe_gradient}
\end{equation}
and therefore introduces a Hessian--vector correction. This curvature is not solely that of the image-space reward. Since $r_i=R_i\circ\phi_i$, the chain rule yields
\begin{equation}
H_i
=
J_{\phi_i}^{\top} H_{R_i}J_{\phi_i}
+
\sum_j
\left[\nabla R_i(\phi_i)\right]_j H_{\phi_{i,j}},
\label{eq:probe_hessian_chain}
\end{equation}
so the correction also captures the local geometry of the condition-specific clean-state prediction. Our default implementation stops gradients through $g_i(x)$ when constructing $x_i$. It consequently uses the first-order approximation $g_i(x_i)$ and avoids the Hessian--vector product in Eq.~\eqref{eq:exact_probe_gradient}, while still recomputing each gradient after its objective-specific intervention. Thus, it is not equivalent to combining $g_{\text{edit}}(x)$ and $g_{\text{fid}}(x)$ at the current state. By evaluating gradients at these perturbed states rather than solely at the current latent, our approach anticipates how hypothetical updates affect the predicted clean endpoints and their associated rewards. This yields a locally non-myopic update without explicitly unrolling future denoising steps.

\paragraph{Relation to explicit denoising look-ahead.}
Let $\Phi^{c}_{t\rightarrow s}$ denote the exact conditional flow map from time $t$ to $s$. Flow composition gives
\begin{equation}
\Phi^{c}_{t\rightarrow 0}(z)
=
\Phi^{c}_{t-h\rightarrow 0}\!\left(
\Phi^{c}_{t\rightarrow t-h}(z)
\right).
\label{eq:flow_composition_lookahead}
\end{equation}
Accordingly, predicting the clean endpoint directly from a probe and first advancing that probe by one denoising step lead to the same endpoint under the exact flow. With the practical velocity estimator and numerical solver, they need not be identical, but they remain close when the local flow is approximately straight, as encouraged by rectified-flow training. We evaluate both realizations on the fixed PIE-Bench ablation subset. Table~\ref{tab:explicit_rollout_ablation} verifies this connection empirically: explicit one-step rollout gives comparable endpoint metrics with mixed metric-wise differences, while increasing runtime by approximately 31\% and peak memory by approximately 16\%. Recomputing the post-intervention clean endpoint therefore captures much of the useful information from one explicit future step at lower cost.

\begin{table*}[t]
\centering
\scriptsize
\caption{Post-intervention clean-endpoint prediction versus explicitly applying one denoising transition before reward evaluation.}
\label{tab:explicit_rollout_ablation}
\resizebox{\textwidth}{!}{
\begin{tabular}{lccccccccc}
\toprule
Method & Structure $\downarrow$ & PSNR $\uparrow$ & LPIPS $\downarrow$ & MSE $\downarrow$ & SSIM $\uparrow$ & CLIP Similarity (Whole) $\uparrow$ & CLIP Similarity (Edited) $\uparrow$ & Runtime / image (s) $\downarrow$ & Peak GPU memory (GiB) $\downarrow$ \\
\midrule
Look-ahead (Ours) & \textbf{0.0091} & 27.2950 & \textbf{0.0751} & 0.0026 & 0.8476 & \textbf{27.2370} & 22.0130 & \textbf{185.88} & \textbf{55.82} \\
Look-ahead with one-step denoising & 0.0114 & \textbf{27.3860} & 0.0757 & \textbf{0.0025} & \textbf{0.8510} & 27.1912 & \textbf{22.0686} & 243.70 & 64.88 \\
\bottomrule
\end{tabular}}
\end{table*}

\subsection{Update-Rule Ablation}
\label{app:update_rule_ablation}

We isolate the contribution of the look-ahead rule from the surrounding constrained framework. First, we evaluate edit-first and look-ahead on the full PIE-Bench using the same inversion-free initialization, rewards, constraints, and hyperparameters; only the proposal rule changes. Table~\ref{tab:update_rule_full} shows that the two rules reach nearly identical endpoint performance under unrestricted correction, with mixed wins across metrics. This indicates that the constrained framework accounts for most of the overall gain in Table~\ref{tab:main_results}.

\begin{table*}[t]
\centering
\small
\caption{Edit-first versus look-ahead on the full PIE-Bench using identical hyperparameters and unrestricted correction.}
\label{tab:update_rule_full}
\resizebox{\textwidth}{!}{
\begin{tabular}{lccccccc}
\toprule
Update rule & Structure $\downarrow$ & PSNR $\uparrow$ & LPIPS $\downarrow$ & MSE $\downarrow$ & SSIM $\uparrow$ & CLIP Similarity (Whole) $\uparrow$ & CLIP Similarity (Edited) $\uparrow$ \\
\midrule
Edit-first & \textbf{0.0175} & \textbf{25.0160} & 0.0965 & \textbf{0.0045} & \textbf{0.8442} & \textbf{28.7452} & 23.7910 \\
Look-ahead & 0.0176 & 25.0050 & \textbf{0.0957} & \textbf{0.0045} & 0.8433 & 28.6262 & \textbf{23.7916} \\
\bottomrule
\end{tabular}}
\end{table*}

We then control correction cost on the fixed PIE-Bench ablation subset. For each image, edit-first receives the corresponding projection budget used by look-ahead and stops correcting once that budget is exhausted. As shown in Table~\ref{tab:update_rule_matched_budget}, look-ahead improves all source-preservation metrics and whole-image target alignment while retaining comparable edited-region alignment. Its main benefit is therefore to reach a favorable operating point with less reliance on corrective projections, rather than to shift the unrestricted endpoint frontier substantially.

\subsection{User Study}
\label{app:user_study}

To complement automatic metrics, we conducted a user study with eight adult participants recruited through Prolific. Participants rated anonymized outputs from seven methods on 20 non-sensitive PIE-Bench examples. Each participant evaluated 21 outputs using two five-point Likert scales: edit compliance and source fidelity. Table~\ref{tab:user_study} shows that our method achieves the highest source-fidelity score while maintaining competitive edit compliance.

Participation was voluntary and restricted to adults, and participants provided electronic informed consent before beginning. They were compensated at \pounds 6 per hour in accordance with Prolific's minimum payment requirement. The study involved rating ordinary, non-sensitive images and posed no risks beyond routine web image viewing; these conditions were disclosed before consent. None of the conditions requiring prior ethical review under the Finnish National Board on Research Integrity (TENK) guidelines applied, so prior ethical review was not required.

\begin{table}[t]
\centering
\small
\caption{User study results on five-point Likert scales.}
\label{tab:user_study}
\begin{tabular}{lcc}
\toprule
Method & Edit Compliance $\uparrow$ & Source Fidelity $\uparrow$ \\
\midrule
DDIM Inversion & 2.83 & 2.00 \\
TABA & 2.92 & 2.21 \\
RF-Inversion & 3.12 & 2.58 \\
RF-Solver-Edit & 3.33 & 3.29 \\
FlowEdit & \textbf{3.46} & 3.21 \\
FlowAlign & 3.04 & 2.83 \\
Ours & 3.33 & \textbf{3.71} \\
\bottomrule
\end{tabular}
\end{table}

\begin{figure*}[t]
  \centering
  \includegraphics[width=0.65\textwidth]{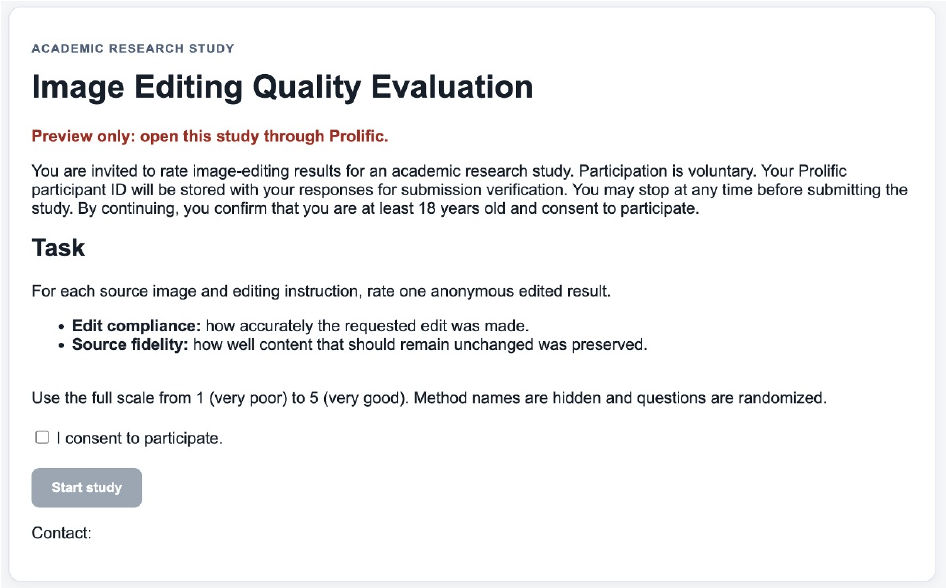}
  \caption{Instructions and consent form shown to participants in the user study.}
  \label{fig:user_study_instruction}
\end{figure*}

\begin{figure*}[t]
  \centering
  \includegraphics[width=0.65\textwidth]{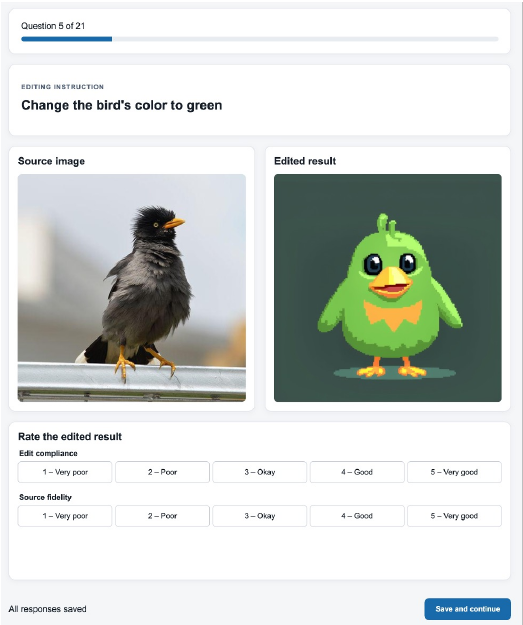}
  \caption{Example rating question from the user study. Method identities were hidden from participants.}
  \label{fig:user_study_case}
\end{figure*}

Figures~\ref{fig:user_study_instruction} and~\ref{fig:user_study_case} show the participant instructions and an example evaluation screen.

\subsection{Computational Cost}
\label{app:computational_cost}
\label{app:compute}

We benchmark our method and the baselines that we run locally on the same NVIDIA H200 GPU and report per-image runtime and peak allocated GPU memory in Table~\ref{tab:runtime_memory}. We include ITOC~\cite{changTrainingFreeRewardGuidedImage2026}, which performs full-horizon trajectory-control optimization, as context for another reward-gradient-based editor. Our method is substantially more expensive than methods with fixed inversion and generation procedures because it repeatedly evaluates both reward gradients and applies fidelity corrections during generation. This overhead is an important practical limitation.

Our default implementation uses a first-order stop-gradient approximation. Exact differentiation through the probe gradients introduces Hessian--vector products and requires 392.0 seconds and 129.76 GiB per image, compared with 185.88 seconds and 55.82 GiB for the first-order implementation, i.e., $2.11\times$ runtime and $2.32\times$ peak memory. Explicitly advancing each probe by one additional denoising step yields comparable endpoint metrics but increases runtime to 243.70 seconds and peak memory to 64.88 GiB. These results motivate post-intervention clean-endpoint prediction without exact second-order differentiation or explicit rollout.

We additionally consider a more efficient variant based on a linearized projection.
Using the look-ahead objective $\mathcal{L}_{\mathrm{LA}}$ defined in Eq.~\eqref{eq:lookahead_objective}, we denote its nominal latent increment by
\begin{equation}
\Delta x_{\mathrm{LA}}
:=
\gamma
\frac{\nabla_x\mathcal{L}_{\mathrm{LA}}(x)}
{\lVert\nabla_x\mathcal{L}_{\mathrm{LA}}(x)\rVert_2}.
\end{equation}
Here, $\Delta x$ denotes an increment applied to the current latent $x$.
We approximate the original fidelity constraint
$r_{\text{fid}}(x+\Delta x,t)\geq\varepsilon$ by its first-order expansion,
\begin{equation}
r_{\text{fid}}(x,t)
+
\nabla_x r_{\text{fid}}(x,t)^\top\Delta x
\geq\varepsilon.
\end{equation}
The projected update is the minimum Euclidean correction to the look-ahead increment:
\begin{equation}
\Delta x^\star
=
\underset{\Delta x}{\arg\min}\;
\frac{1}{2}
\lVert\Delta x-\Delta x_{\mathrm{LA}}\rVert_2^2
\quad
\text{s.t.}\quad
r_{\text{fid}}(x,t)
+
\nabla_x r_{\text{fid}}(x,t)^\top\Delta x
\geq\varepsilon.
\end{equation}
Its Karush--Kuhn--Tucker solution is
\begin{equation}
\Delta x^\star
=
\Delta x_{\mathrm{LA}}
+
\mu^\star\nabla_x r_{\text{fid}}(x,t),
\qquad
\mu^\star
=
\left[
\frac{
\varepsilon-r_{\text{fid}}(x,t)
-\nabla_x r_{\text{fid}}(x,t)^\top\Delta x_{\mathrm{LA}}
}{
\lVert\nabla_x r_{\text{fid}}(x,t)\rVert_2^2
}
\right]_+,
\end{equation}
where $\mu^\star$ is the non-negative Lagrange multiplier and
$[z]_+:=\max(z,0)$ denotes the positive-part operator.
We then update the latent as $x\leftarrow x+\Delta x^\star$.
Thus, the look-ahead increment is unchanged when it satisfies the local fidelity constraint and otherwise receives the smallest first-order feasibility correction.
This closed-form projection replaces iterative trust-region corrections and reduces the runtime from 185.9 to 139.4 seconds per image, corresponding to a $1.33\times$ speedup.

\subsection{Sensitivity and Design Ablations}
\label{app:sensitivity_ablations}

Unless noted otherwise, these ablations use the fixed PIE-Bench ablation subset and the same evaluation protocol. They vary one component at a time while keeping all remaining settings fixed.

\paragraph{Inner iterations and bounded correction.}
Table~\ref{tab:p_j_sensitivity} reports sensitivity to the number of inner updates $p$ and maximum correction steps $J$. Increasing $p$ from 1 to 3 strengthens editing and improves several preservation metrics, whereas $p=4$ provides only mixed marginal changes at additional cost. Increasing $J$ improves approximate feasibility and preservation up to $J=3$; moving to $J=5$ increases the average number of corrections by roughly 46\% with only marginal metric changes. The finite loop is therefore a computation-bounded approximation: if the constraint remains violated after $J$ corrections, sampling continues from the last corrected state rather than guaranteeing exact feasibility.

\begin{table*}[t]
\centering
\scriptsize
\caption{Sensitivity to inner updates $p$ and maximum correction steps $J$. Dashes denote quantities not applicable to the corresponding sweep.}
\label{tab:p_j_sensitivity}
\resizebox{\textwidth}{!}{
\begin{tabular}{llccccccccc}
\toprule
Sweep & Value & Structure $\downarrow$ & PSNR $\uparrow$ & LPIPS $\downarrow$ & MSE $\downarrow$ & SSIM $\uparrow$ & CLIP Whole $\uparrow$ & CLIP Edited $\uparrow$ & Violations/image $\downarrow$ & Updates/image \\
\midrule
\multirow{4}{*}{$p$} & 1 & 0.0126 & \textbf{27.4450} & 0.0795 & 0.0025 & 0.8454 & 25.9090 & 21.1890 & -- & 28.00 \\
 & 2 & 0.0101 & 27.4370 & 0.0770 & 0.0025 & \textbf{0.8522} & 25.7660 & 21.2480 & -- & 56.00 \\
 & 3 & \textbf{0.0091} & 27.2950 & \textbf{0.0751} & 0.0026 & 0.8476 & \textbf{27.2370} & 22.0130 & -- & 84.00 \\
 & 4 & 0.0103 & 27.4060 & 0.0752 & \textbf{0.0024} & 0.8508 & 27.1870 & \textbf{22.1060} & -- & 112.00 \\
\midrule
\multirow{4}{*}{$J$} & 1 & 0.0129 & 27.2300 & 0.0835 & 0.0026 & 0.8409 & 26.4750 & 21.1020 & 64.22 & -- \\
 & 2 & 0.0101 & 27.2550 & 0.0781 & \textbf{0.0025} & 0.8456 & \textbf{27.9370} & 21.6040 & 55.61 & 103.56 \\
 & 3 & \textbf{0.0091} & \textbf{27.2950} & \textbf{0.0751} & 0.0026 & 0.8476 & 27.2370 & 22.0130 & 52.56 & 139.94 \\
 & 5 & 0.0094 & 27.2680 & 0.0755 & 0.0026 & \textbf{0.8529} & 27.0150 & \textbf{22.4090} & \textbf{48.67} & 203.94 \\
\bottomrule
\end{tabular}}
\end{table*}

\paragraph{Look-ahead weighting and probe distances.}
The weight $\lambda$ acts on objective-specific post-intervention rewards, while feasibility is controlled separately by $\varepsilon$. Table~\ref{tab:lookahead_parameter_sensitivity} shows that preservation metrics vary modestly over a $16\times$ range of $\lambda$. We use $\lambda=1$ without claiming that it optimizes every metric. We use the shared-step convention $\alpha=\beta=\gamma=5$ with normalized gradients. Varying either probe distance from 1 to 10 produces a stable preservation profile and no uniformly optimal setting. In contrast, direct scalarization controls feasibility only indirectly through reward scales; its empirical edit--fidelity curve is dominated by the constrained method, particularly under strict fidelity requirements.

\begin{table*}[t]
\centering
\scriptsize
\caption{Sensitivity to fidelity weight $\lambda$, edit-probe distance $\alpha$, and fidelity-probe distance $\beta$.}
\label{tab:lookahead_parameter_sensitivity}
\resizebox{\textwidth}{!}{
\begin{tabular}{llccccccc}
\toprule
Sweep & Value & Structure $\downarrow$ & PSNR $\uparrow$ & LPIPS $\downarrow$ & MSE $\downarrow$ & SSIM $\uparrow$ & CLIP Whole $\uparrow$ & CLIP Edited $\uparrow$ \\
\midrule
\multirow{5}{*}{$\lambda$} & 0.25 & 0.0094 & 27.3820 & 0.0769 & 0.0025 & 0.8498 & 26.9050 & 21.9750 \\
 & 0.5 & 0.0094 & 27.2030 & 0.0745 & 0.0026 & 0.8471 & 27.1840 & 21.7970 \\
 & 1.0 & \textbf{0.0091} & 27.2950 & 0.0751 & 0.0026 & 0.8476 & \textbf{27.2370} & 22.0130 \\
 & 2.0 & 0.0105 & 27.3680 & 0.0771 & 0.0025 & \textbf{0.8519} & 26.7390 & \textbf{22.4200} \\
 & 4.0 & 0.0103 & \textbf{27.5620} & \textbf{0.0710} & \textbf{0.0024} & 0.8518 & 26.5090 & 21.2290 \\
\midrule
\multirow{5}{*}{$\alpha$} & 1.0 & 0.0095 & 27.2720 & 0.0766 & 0.0026 & 0.8508 & 26.8310 & 21.8640 \\
 & 2.5 & 0.0094 & 27.3070 & 0.0762 & 0.0026 & 0.8493 & 27.0360 & \textbf{22.6770} \\
 & 5.0 & \textbf{0.0091} & 27.2950 & \textbf{0.0751} & 0.0026 & 0.8476 & 27.2370 & 22.0130 \\
 & 7.5 & 0.0096 & 27.2650 & 0.0799 & 0.0026 & 0.8492 & 27.3760 & 22.6180 \\
 & 10.0 & 0.0097 & \textbf{27.3380} & \textbf{0.0751} & \textbf{0.0025} & \textbf{0.8529} & \textbf{27.6090} & 22.1660 \\
\midrule
\multirow{5}{*}{$\beta$} & 1.0 & 0.0101 & \textbf{27.4360} & 0.0764 & \textbf{0.0025} & \textbf{0.8532} & 27.4960 & 21.7550 \\
 & 2.5 & 0.0096 & 27.4050 & 0.0779 & \textbf{0.0025} & 0.8527 & 27.3590 & \textbf{22.5680} \\
 & 5.0 & \textbf{0.0091} & 27.2950 & \textbf{0.0751} & 0.0026 & 0.8476 & 27.2370 & 22.0130 \\
 & 7.5 & 0.0097 & 27.2680 & 0.0759 & 0.0026 & 0.8501 & 26.9560 & 21.5660 \\
 & 10.0 & 0.0095 & 27.1570 & 0.0781 & 0.0026 & 0.8503 & \textbf{28.3000} & 22.0030 \\
\bottomrule
\end{tabular}}
\end{table*}

\paragraph{Reward and initialization choices.}
Table~\ref{tab:reward_initialization_ablation} compares the edit rewards described in Appendix~\ref{sec:edit_reward} and the two initialization strategies. Text-image similarity attains the highest target CLIP but remains close to the source prompt; contrastive similarity produces the largest target--source margin and best structure distance. Inversion-free initialization improves every preservation metric and edited-region alignment, with comparable whole-image target alignment.

\begin{table*}[t]
\centering
\scriptsize
\caption{Edit-reward and initialization ablations. Source/target CLIP and margin are reported only for the reward comparison.}
\label{tab:reward_initialization_ablation}
\resizebox{\textwidth}{!}{
\begin{tabular}{llcccccccccc}
\toprule
Type & Variant & Structure $\downarrow$ & PSNR $\uparrow$ & LPIPS $\downarrow$ & MSE $\downarrow$ & SSIM $\uparrow$ & Source CLIP & Target CLIP $\uparrow$ & CLIP Edited $\uparrow$ & Margin $\uparrow$ \\
\midrule
\multirow{3}{*}{Edit reward} & Text-image & 0.0100 & 27.4210 & 0.0765 & \textbf{0.0025} & \textbf{0.8518} & 34.4100 & \textbf{37.1330} & \textbf{26.8110} & 2.7230 \\
 & Directional & 0.0094 & \textbf{27.5340} & \textbf{0.0749} & \textbf{0.0025} & 0.8507 & 22.3930 & 27.2300 & 22.4150 & 4.8370 \\
 & Contrastive & \textbf{0.0091} & 27.2950 & 0.0751 & 0.0026 & 0.8476 & 18.7140 & 27.2370 & 22.0130 & \textbf{8.5240} \\
\midrule
\multirow{2}{*}{Initialization} & Inversion-free & \textbf{0.0091} & \textbf{27.2950} & \textbf{0.0751} & \textbf{0.0026} & \textbf{0.8476} & -- & 27.2370 & \textbf{22.0130} & -- \\
 & Inversion & 0.0155 & 25.5560 & 0.0845 & 0.0038 & 0.8385 & -- & \textbf{27.5780} & 21.9290 & -- \\
\bottomrule
\end{tabular}}
\end{table*}

\paragraph{Alternative differentiable guidance rewards.}
Because using CLIP for both optimization and evaluation can introduce circularity, we replace CLIP-contrastive guidance with ImageReward and HPS v2.1 while keeping the constrained procedure and all other hyperparameters fixed. Table~\ref{tab:guidance_reward_ablation} shows comparable preservation and edited-region CLIP alignment under both non-CLIPScore objectives. The framework is therefore not tied to CLIP-contrastive guidance.

\begin{table}[t]
\centering
\scriptsize
\caption{Guidance-reward replacement on the PIE-Bench ablation subset.}
\label{tab:guidance_reward_ablation}
\resizebox{\columnwidth}{!}{
\begin{tabular}{lccccccc}
\toprule
Reward & Structure $\downarrow$ & PSNR $\uparrow$ & LPIPS $\downarrow$ & MSE $\downarrow$ & SSIM $\uparrow$ & CLIP Whole $\uparrow$ & CLIP Edited $\uparrow$ \\
\midrule
CLIP-contrastive & \textbf{0.0091} & \textbf{27.2950} & \textbf{0.0751} & 0.00260 & \textbf{0.8476} & \textbf{27.2370} & 22.0130 \\
ImageReward & 0.0125 & 27.1130 & 0.0777 & \textbf{0.00256} & 0.8475 & 26.1243 & 21.9768 \\
HPS v2.1 & 0.0124 & 26.9636 & 0.0776 & 0.00274 & 0.8471 & 27.1997 & \textbf{22.4951} \\
\bottomrule
\end{tabular}}
\end{table}

\paragraph{Fidelity reward.}
A DINO-only constraint improves edited-region alignment slightly but severely degrades low-level source preservation, as semantic features are less sensitive to exact color, texture, and local geometry. Combining DINO with latent-$\ell_2$ can improve individual fidelity metrics, but introduces another weight-dependent trade-off and does not uniformly dominate latent-$\ell_2$. For example, weight 0.020 improves LPIPS from 0.0751 to 0.0736 but reduces edited-region CLIP from 22.0130 to 20.6478. We therefore retain latent-$\ell_2$ as the default and report the full sensitivity in Table~\ref{tab:fidelity_reward_ablation}.

\begin{table*}[t]
\centering
\scriptsize
\caption{Fidelity-objective ablation at threshold $\varepsilon=-0.1$. The weight applies to the DINO term in the combined objective.}
\label{tab:fidelity_reward_ablation}
\resizebox{\textwidth}{!}{
\begin{tabular}{llccccccc}
\toprule
Objective & Weight & Structure $\downarrow$ & PSNR $\uparrow$ & LPIPS $\downarrow$ & MSE $\downarrow$ & SSIM $\uparrow$ & CLIP Whole $\uparrow$ & CLIP Edited $\uparrow$ \\
\midrule
Latent-$\ell_2$ & 0 & \textbf{0.0091} & 27.2950 & 0.0751 & 0.0026 & 0.8476 & 27.2370 & 22.0130 \\
DINO & -- & 0.2106 & 9.6213 & 0.4456 & 0.1427 & 0.5122 & 26.8204 & \textbf{22.7671} \\
Latent-$\ell_2$ + DINO & 0.005 & 0.0118 & 26.9860 & 0.0862 & 0.0028 & 0.8478 & 26.5615 & 21.2238 \\
 & 0.020 & 0.0096 & \textbf{27.4742} & \textbf{0.0736} & \textbf{0.0024} & \textbf{0.8548} & 26.6459 & 20.6478 \\
 & 0.050 & 0.0173 & 26.6272 & 0.0952 & 0.0036 & 0.8392 & 25.7457 & 21.4099 \\
 & 0.100 & 0.0154 & 27.0727 & 0.0851 & 0.0029 & 0.8453 & 26.9130 & 21.3573 \\
 & 0.200 & 0.0310 & 25.2286 & 0.1138 & 0.0061 & 0.8030 & 26.4817 & 21.5588 \\
 & 0.300 & 0.0280 & 25.5062 & 0.1032 & 0.0043 & 0.8060 & \textbf{27.5068} & 22.0360 \\
\bottomrule
\end{tabular}}
\end{table*}

\section{Qualitative Effect of Fidelity Threshold}
\label{app:eps_qualitative}

\begin{figure}[h]
  \centering
  \includegraphics[width=\linewidth]{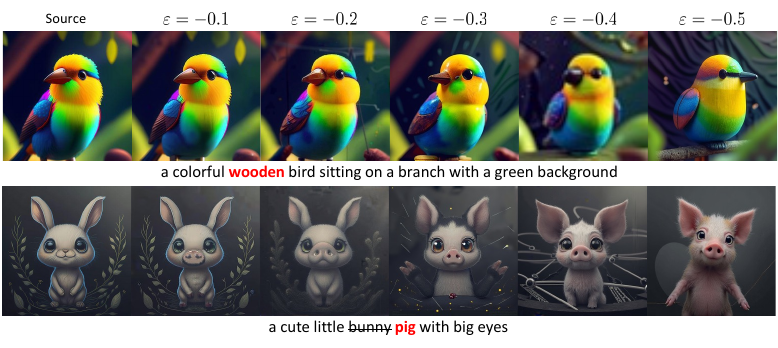}
  \caption{
    \textbf{Qualitative effect of varying $\varepsilon$.}
    Each row shows the source image followed by results under $\varepsilon \in \{-0.1, -0.2, -0.3, -0.4, -0.5\}$.
    Looser constraints produce stronger edits with greater source deviation.
  }
  \label{fig:eps_qualitative}
\end{figure}

Figure~\ref{fig:eps_qualitative} illustrates how the fidelity threshold $\varepsilon$ controls the edit strength on two representative examples.
As $\varepsilon$ becomes more negative (looser constraint), the edit progressively strengthens: in the first case, the bird's texture gradually transitions toward a wooden appearance while the overall composition is retained; in the second case, the subject shifts continuously from a bunny toward a pig, with intermediate values producing a plausible blend between the two.
At tight $\varepsilon$ (e.g., $\varepsilon = -0.1$), the result closely preserves the source while making only subtle edits; at loose $\varepsilon$ (e.g., $\varepsilon = -0.5$), the edit is strongly executed at the cost of greater source deviation.
This monotonic and smooth behavior confirms that $\varepsilon$ provides an continuous control over the edit--fidelity trade-off.

\section{Failure Cases and Limitations}
\label{app:limitations}

\begin{figure}[H]
  \centering
  \includegraphics[width=\linewidth]{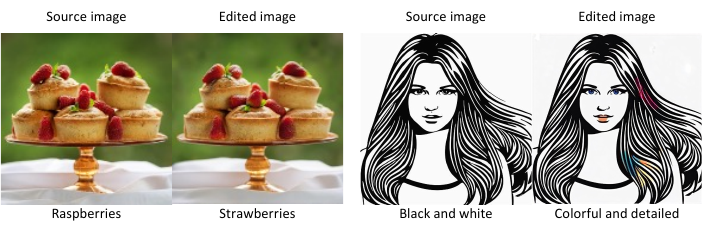}
  \caption{Failure cases. Our method produces results that are too similar to the source when the requested edit requires a large semantic or style change.}
  \label{fig:failure}
\end{figure}

\paragraph{Failure cases.}
Our method tends to complete edits through conservative latent updates, which improves consistency with the source image.
However, when the requested edit requires a large semantic or style change, this conservative tendency leads to insufficient edit strength, producing results that remain overly similar to the source.
Figure~\ref{fig:failure} illustrates two representative cases: changing raspberries to strawberries on cupcakes, and converting a black-and-white line drawing to a colorful and detailed image---both of which are under-edited by our method.

\paragraph{Limitations.}
First, the method introduces repeated reward-gradient evaluations and fidelity corrections at every generation step. It is therefore substantially slower and more memory intensive than editing methods with fixed inversion and generation procedures, as quantified in Appendix~\ref{app:computational_cost}; improving inference efficiency remains important future work.
Second, a strict fidelity threshold can reject updates required for large geometric or semantic changes. Although relaxing $\varepsilon$ provides an explicit control, the method cannot eliminate the fundamental tension between editability and source preservation and may under-edit demanding instructions.
Third, the fidelity reward $r_{\text{fid}}$ is a global measure; it does not enforce region-specific preservation, so edits can occasionally leak into areas that should remain unchanged.
Finally, performance depends on reward quality: if the edit reward does not capture complex relational or compositional instructions, optimization may not converge to the desired output. Alternative differentiable rewards are compatible with the framework, but their own biases remain.

\section{Broader Impacts}
\label{app:broader_impacts}

Our method improves controllability in text-guided image editing, which can benefit creative, educational, and accessibility applications. At the same time, like any generative editing technique, it could be misused to produce deceptive imagery. Our method relies on FLUX.1-dev ~\cite{labs2025flux}under a non-commercial research license that carries its own content restrictions, and standard generative model safeguards apply equally here.

\end{document}